\documentclass[journal]{IEEEtran}
\usepackage{times}
\usepackage{multicol}

\usepackage{xparse} 
\usepackage{amsmath} 
\usepackage{amssymb}
\usepackage{amsfonts} 
\usepackage{dsfont}
\usepackage{graphicx}
\usepackage{url}
\usepackage{booktabs}
\usepackage[para]{threeparttable}
\usepackage{multirow}
\usepackage{makecell}
\usepackage{tikz}

\usepackage{amsthm}
\usepackage[ruled,vlined]{algorithm2e}
\usepackage{cite} 
\usepackage[colorlinks,bookmarksopen,bookmarksnumbered,citecolor=red,urlcolor=red]{hyperref}
\usepackage[capitalize]{cleveref}
\usepackage{xcolor}
\usepackage{framed}
\usepackage{caption}
\usepackage{subcaption}

\SetKwInOut{Parameters}{Parameters}

\usepackage{setspace}

\colorlet{shadecolor}{gray!10}

\NewDocumentCommand\bbm{}{ \begin{bmatrix} }
\NewDocumentCommand\ebm{}{ \end{bmatrix} }
\NewDocumentCommand\Vector{m}{ \boldsymbol{\mathbf{#1}} }
\NewDocumentCommand\Matrix{m}{ \boldsymbol{\mathbf{#1}} }
\NewDocumentCommand\Transpose{m}{ \left.{#1}\right.^{\! T} }

\NewDocumentCommand\Norm{m}{\left\Vert#1\right\Vert }

\NewDocumentCommand\Real{}{ \mathbb{R} }

\NewDocumentCommand\LieGroupSO{m}{ \mathrm{SO}(#1) }

\NewDocumentCommand\LieGroupSE{m}{ \mathrm{SE}(#1) }

\NewDocumentCommand\NormalDistribution{mm}{ \mathcal{N}\left(#1,#2\right) }

\NewDocumentCommand\Identity{}{ \Matrix{I} }

\NewDocumentCommand\CoordinateFrame{m}{ \underrightarrow{\Matrix{\mathcal{F}}}_{#1} }

\begin{document}
\title{Spatiotemporal Calibration of\\ 3D Millimetre-Wavelength Radar-Camera Pairs}
\author{Emmett Wise, Qilong Cheng, and Jonathan Kelly
\thanks{Emmett Wise, Qilong Cheng, and Jonathan Kelly are with the Space and Terrestrial Autonomous Robotic Systems Laboratory, University of Toronto, Institute for Aerospace Studies, Toronto, Canada. \{\texttt{<first name>.<last name>@robotics.utias.utoronto.ca}\}}
\thanks{This work was supported in part by the Natural Sciences and Engineering Research Council of Canada (NSERC).
Jonathan Kelly was supported by the Canada Research Chairs Program.
Jonathan Kelly is a Vector Institute for Artificial Intelligence Faculty Affiliate.}
}

\newenvironment{rcases}
  {\left.\begin{aligned}}
  {\end{aligned}\right\rbrace}
\newcommand{\qc}[1]{\textcolor{purple}{\textbf{{QC:} #1}}}
\newcommand{\ew}[1]{\textcolor{orange}{\textbf{{EW:} #1}}}
\newcommand{\jk}[1]{\textcolor{red}{\textbf{{JK:} #1}}}
\newcommand{\edit}[1]{\textcolor{blue}{#1}}

\markboth{IEEE Transactions on Robotics}%
{Shell \MakeLowercase{\textit{et al.}}: Bare Demo of IEEEtran.cls for IEEE Journals}

\maketitle

\begin{abstract}
Autonomous vehicles (AVs) often depend on multiple sensors and sensing modalities to impart a measure of robustness when operating in adverse conditions.
Radars and cameras are popular choices for use in combination; although radar measurements are sparse in comparison to camera images, radar scans are able to penetrate fog, rain, and snow.
Data from both sensors are typically fused prior to use in downstream perception tasks.
However, accurate sensor fusion depends upon knowledge of the spatial transform between the sensors and any temporal misalignment that exists in their measurement times.
During the life cycle of an AV, these calibration parameters may change, so the ability to perform in-situ spatiotemporal calibration is essential to ensure reliable long-term operation.
State-of-the-art 3D radar-camera spatiotemporal calibration algorithms require bespoke calibration targets that are not readily available in the field.  
In this paper, we describe an algorithm for \emph{targetless} spatiotemporal calibration that is able to operate without specialized infrastructure.
Our approach leverages the ability of the radar unit to measure its own ego-velocity relative to a fixed, external reference frame.
We analyze the identifiability of the spatiotemporal calibration problem and determine the motions necessary for calibration.
Through a series of simulation studies, we characterize the sensitivity of our algorithm to measurement noise.
Finally, we demonstrate accurate calibration for three real-world systems, including a handheld sensor rig and a vehicle-mounted sensor array.
Our results show that we are able to match the performance of an existing, target-based method, while calibrating in arbitrary, infrastructure-free environments. 
\end{abstract}


\IEEEpeerreviewmaketitle

\section{Introduction}
\label{sec:intro}

The widespread deployment of autonomous vehicles (AVs) depends critically on their ability to operate safely under a range of challenging environmental conditions.
To ensure sufficient redundancy, most AV perception systems incorporate multiple sensors and sensing modalities.
In this paper, we consider 3D mm-wavelength radar as a complementary sensor to standard cameras for safe AV perception.

The operating principle of mm-wavelength radars (i.e., the active emission of mm-wavelength electromagnetic (EM) radiation) makes these sensors relatively immune to adverse conditions that negatively affect cameras.
Radars also provide information that cameras do not, including \emph{range-rate} measurements of the relative velocity of targets in the environment. 
However, radar data are much lower resolution and significantly more noisy, than visual measurements under nominal conditions.
Together, radars and cameras are highly complementary, providing situational awareness under both nominal and visually-degraded conditions.

\begin{figure}[t]
	\centering
	\vspace{2mm}
	\begin{tikzpicture}
	\node[anchor=south west,inner sep=0] (image) at (0,0) {\includegraphics[width=0.85\columnwidth]{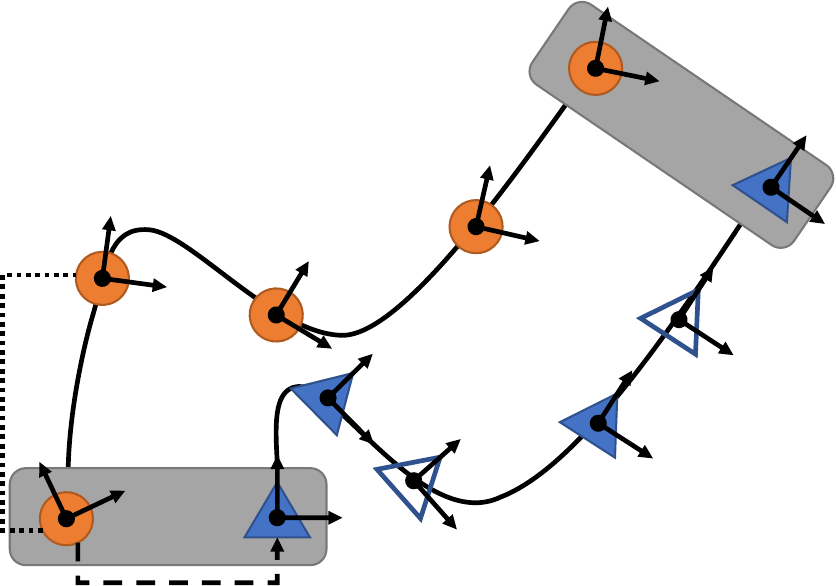}};
	\draw (-0.3,0.0) node {Camera};
	\draw (3.25, 0.0) node {Radar};
	\node (traj) [align=center] at (1.7,3.8) {Continuous-Time\\Trajectory};
	\draw (1.65,-0.35) node {$\Matrix{T}_{cr}$};
	\draw (-0.3,1.7) node {$\alpha$};
	\draw (0.7,3.2) node {$t_k$};
	\draw (3.4,1.75) node {$t_{j}$};
	\draw (4.65,0.45) node {$t_{j} + \tau$};
	\draw (3.1,2.65) node {$t_{k+1}$};
	\draw (5.2,1.9) node {$t_{j+1}$};
	\draw (6.75,1.85) node {$t_{j+1} + \tau$};
	\draw (4.95,3.45) node {$t_{k+2}$};
	\end{tikzpicture}
	\caption{The radar (triangle) and camera (circle) are assumed to be rigidly connected. Our calibration problem involves estimating the transform between the camera and radar, $\Matrix{T}_{cr}$, the translation scale factor, $\alpha$, for the camera pose measurements, and the temporal offset, $\tau$. The unfilled triangles represent radar measurements at “shifted” points in time due to the offset bias, which must be considered to ensure the correct radar ego-velocity estimate.}
	\label{fig:diagram}
\end{figure}
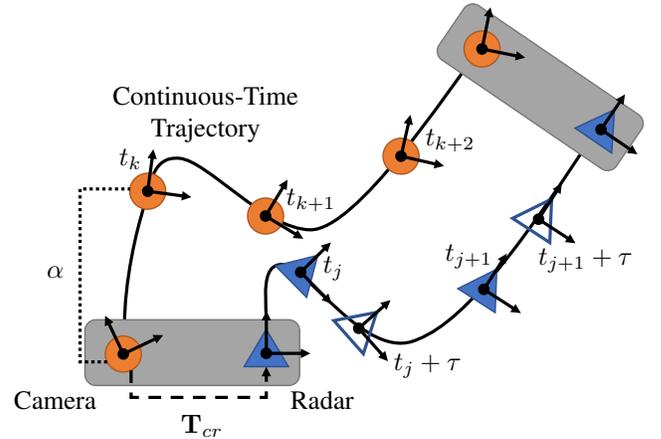

To be used jointly in the AV perception stack, radar and camera sensors must be calibrated with respect to each other. 
The spatial (6-DoF) transform between a radar-camera pair must be known accurately in order to express their data in a common reference frame.
An AV may undergo calibration `at the factory' prior to operation, but maintenance and general wear and tear can alter the spatial calibration parameters.
Further performance gains are achieved when the sensor data streams are temporally aligned in addition to spatial alignment.
Even when the sensors are externally triggered, internal signal processing delays can result in shifted measurement timestamps.
If this time offset is not accounted for, then, for example, moving targets will be \emph{spatially} shifted in the radar and camera data.
Further, in some systems, power cycling or reconfiguring the sensors may change the time offset.
In turn, temporal calibration may need to be performed routinely to ensure the accuracy and integrity of fused sensor measurements.

An in-situ method to estimate the spatial transformation from the radar to the camera and the temporal offset of the sensor data streams, would enable long-term AV operation in the field.
However, existing radar-camera spatial and spatiotemporal calibration algorithms are restricted to certain environments and sensor configurations \cite{lee_extrinsic_2020,persic_spatiotemporal_2021}.
Primarily, these methods rely on the assumption that the radar measures `point-like' reflections from objects. 
In general, a radar measurement, determined from a reflected EM pulse, is a complex function of the shape, relative orientation, size, and composition of an object \cite{richards_principle_2010}.
Additionally, multipath reflections introduce outlier measurements of ghost `objects' \cite{richards_principle_2010}.
To avoid these problems, specialized trihedral retroreflective radar targets are used to produce the desired point-like radar returns. 
A visual fiducial can be placed over or alongside a trihedral target, allowing radar-camera measurement correspondences to be established.
The use of targets, however, means that calibration must be carried out in specialized areas or with infrastructure that is not usually available during regular AV operation.
Additionally, these algorithms require that the radar-camera pair(s) share overlapping fields of view, which may not be possible for all radar-camera systems.

Herein, we extend the method in \cite{wise_continuous-time_2021} to jointly estimate the extrinsic calibration parameters and temporal offset of a 3D mm-wavelength radar-camera pair in a fully targetless manner.
Importantly, our approach does not require the sensors to share overlapping fields of view.
Instead, we use radar measurements to estimate the instantaneous radar ego-velocity --- that is, the velocity of the radar unit relative to an external reference frame expressed in the radar reference frame \cite{Stahoviak_Velocity_2019}.
By relying on velocity information, we remove the need for specialized calibration targets while also avoiding the difficult problems of radar and cross-modal data association.
We make the following contributions:
\begin{itemize}
	\item we extend the work in \cite{wise_continuous-time_2021} to enable full spatiotemporal calibration of monocular camera-3D radar pairs in arbitrary configurations;
	\item we prove that the calibration problem is identifiable and determine the motions that are required for reliable calibration;
	\item we analyze the accuracy of spatiotemporal calibration with varying amounts of sensor noise through an extensive series of simulation studies; and
	\item we carry out three different real-world experiments, which demonstrate that our algorithm is able to match the accuracy of an existing, target-based method and that we are able to perform calibration in different environments, including for sensors on board an AV.
\end{itemize}

In \Cref{sec:related}, we survey existing extrinsic and spatiotemporal calibration algorithms for mm-wavelength radar sensors.
\Cref{sec:methodology} formulates spatiotemporal calibration as a batch, continuous-time estimation problem.
We examine the identifiability of the calibration problem in \Cref{sec:ident}.
In \Cref{sec:sim-experiments}, we describe two simulation experiments designed to evaluate the robustness of our algorithm.
In \Cref{sec:real-experiments}, we demonstrate the accuracy and flexibility of our algorithm by reporting on three real-world experiments in different environments.
Finally, we summarize our work in \Cref{sec:conclusion}.

\section{Related Work}
\label{sec:related}

In this section, we survey spatial and spatiotemporal calibration algorithms that can be applied to mm-wavelength radars in relation to another, complementary sensor.
\Cref{sec:ext-cal-target} reviews algorithms for target-based extrinsic calibration, while \Cref{sec:ext-cal} describes algorithms for target-free, or \emph{targetless}, extrinsic calibration.
In \Cref{sec:target-spatiotemporal}, we discuss prior work on target-based spatiotemporal calibration.

\subsection{Target-Based Extrinsic Calibration}
\label{sec:ext-cal-target}

Early radar extrinsic calibration algorithms, developed prior to the widespread availability of 3D mm-wavelength radar units, were designed to enable 2D radar-camera data fusion.
Many of these early extrinsic calibration techniques operate by computing the projective homography that maps points on the horizontal (sensing) radar plane to points on the camera image plane.
Because radar sensors are inherently noisy, most calibration algorithms require specialized trihedral reflectors, shown in \cref{fig:target}, that produce coincident, point-like `signals' in both the radar and camera data, making the correspondence problem easier to solve \cite{sugimoto_obstacle_2004,Wang2011,kim_data_2014,kim_radar_2018}.
Although 2D radar sensors are not able to properly measure the elevation of remote targets, they do detect targets at a small elevation angle above the radar horizontal plane.
Since the distance to off-plane targets will be slightly different, accurate calibration requires that detected reflectors \textit{do} lie on the radar horizontal plane.
Sugimoto et al.\ \cite{sugimoto_obstacle_2004} constrain the trihedral reflector position using the radar return signal strength.
During calibration, the approach in \cite{sugimoto_obstacle_2004} filters radar-camera measurement pairs by return intensity; the intensity is maximal for reflectors that lie on the horizontal plane.

More recent 2D radar extrinsic calibration algorithms often minimize a type of `reprojection error,' that is, the error in the alignment of identifiable objects that appear within both sensors' fields of view.
Kim et al.\ \cite{kim_comparative_2017} leverage reprojection error to estimate the radar-to-camera transform but assume that radar measurements are strictly constrained to the zero-elevation plane.
El Natour et al.\ \cite{elnatour_radar_2015} determine the radar-to-camera transform by intersecting backprojected camera rays with the 3D `arcs' along which the 2D radar measurements must lie.
Domhof et al.\ \cite{Domhof2019_calibration} use a specialized calibration target that provides scale for the camera measurement, enabling extrinsic calibration via point cloud alignment.
Per\v{s}i\'{c} et al.\ \cite{Persic2019_calibration} also perform extrinsic calibration via 3D point cloud alignment but improve overall accuracy by modelling the relationship between target return intensity and elevation angle.
The `homography' and `reprojection' methods are summarized and compared by Oh et al.\ in \cite{Oh2018_calibration}, where the authors conclude that both have similar performance.
Due to the infrastructure needs (i.e., specialized targets), the methods above are restricted to sensor pairs that share overlapping fields of view. This requirement may be impossible to satisfy for certain sensor configurations.
By leveraging constraints induced by the motion of a rigidly-connected radar-camera pair, we are able to calibrate sensors that do not share overlapping fields of view.
Further, our approach operates without added infrastructure, enabling calibration under a wider range of conditions.

\subsection{Target-Free Extrinsic Calibration}
\label{sec:ext-cal}

Some extrinsic calibration algorithms do not require specialized retroreflective targets.
Sch{\"{o}}ller et al.\ \cite{Knoll2019_calibration} train a neural network end-to-end  to regress a rotation correction from raw camera images and radar data, for example.
Per\v{s}i\'{c} et al.\ \cite{Persic2020} estimate the yaw angles (only) between radar, camera, and lidar sensors by aligning the trajectories of tracked objects.
Both of these methods require manual measurement of the translation parameters and overlapping fields of view.

Heng \cite{heng_automatic_2020} presents the first reprojection error-based 3D radar-lidar extrinsic calibration algorithm that does not require specialized targets or overlapping sensor fields of view. 
The approach in \cite{heng_automatic_2020} estimates the extrinsic calibration between several lidar units and, using a known vehicle trajectory, constructs a 3D point cloud map.
The radar-lidar extrinsic calibration parameters are then determined by minimizing two weighted residuals: i) the distance from the radar point measurements to the closest plane in the lidar map and ii) the radial velocity error.
However, this method requires the construction of a dense lidar map and known vehicle poses.

Instead of using feature positions, a subset of extrinsic calibration algorithms fuse ego-velocity and ego-motion measurements from the radar and second sensor, respectively.
Since the motion of each sensor is estimated separately, these methods do not perform radar or cross-modal data association and are inherently `target-free.'
Kellner et al.\ \cite{Kellner2015_calibration} estimate the rotation between a car-mounted 2D radar and an inertial measurement unit (IMU) by minimizing the difference in estimated lateral velocities expressed in the radar frame.
While the radar ego-velocity measurements provide lateral velocity directly, determining the lateral velocity of the radar from IMU measurements requires both the IMU angular velocity and accurate knowledge of the radar-IMU translation.
Doer et al.\ \cite{doer_radar_2020} extend the approach in \cite{Kellner2015_calibration} to estimate the full extrinsic calibration for a 3D radar-IMU pair.
Their method is able to achieve a spatial calibration accuracy of 5 cm and 5$^\circ$ when using simulated, low-noise (our designation, see \Cref{sec:sim-experiments}) radar measurements. 
Wise et al.\ \cite{wise_continuous-time_2021} perform extrinsic calibration in continuous time using instantaneous radar ego-velocity measurements and camera egomotion measurements.
Given an unknown but fixed temporal offset, the spatial calibration parameters estimated by this method are within 3 cm and 1$^\circ$, per axis, of those determined by \cite{persic_spatiotemporal_2021}.
All of these techniques rely on ad-hoc temporal calibration schemes.
Herein, we incorporate a principled temporal calibration method.

\subsection{Target-Based Spatiotemporal Calibration}
\label{sec:target-spatiotemporal}

To date, only two radar spatiotemporal calibration algorithms have appeared in the literature, by Lee et al.\ \cite{lee_extrinsic_2020} and by Per\v{s}i\'{c} et al.\ \cite{persic_spatiotemporal_2021}.
The algorithm in \cite{lee_extrinsic_2020} first calibrates the 2D radar-lidar spatial transform using the method of Per\v{s}i\'{c} et al.\ \cite{Persic2019_calibration}. 
As a second step, the lidar measurements are expressed in the radar reference frame, and the azimuth error to distant targets is minimized to determine the temporal offset between the sensor data streams.
Per\v{s}i\'{c} et al.\ \cite{persic_spatiotemporal_2021} represent the trajectory of a target moving through the fields of view of multiple sensors using a continuous-time Gaussian process model. 
This representation allows their algorithm to estimate the spatiotemporal calibration parameters by aligning the sensors' trajectories.
In general, jointly determining all parameters as part of one maximum likelihood estimation problem yields superior accuracy \cite{2016_Rehder_General,persic_spatiotemporal_2021}.
Notably, since the methods in \cite{lee_extrinsic_2020} and \cite{persic_spatiotemporal_2021} rely on known targets, they have the same limitations as the methods discussed in \Cref{sec:ext-cal-target}.

\section{Methodology}
\label{sec:methodology}

We formulate radar-to-camera spatiotemporal calibration as a continuous-time batch estimation problem.
In \Cref{sec:notation}, we describe the mathematical notation used throughout the paper.
We choose to parameterize the smooth radar-camera trajectories using continuous-time B-splines; we review the properties of this representation in \Cref{sec:b-splines}.
In \Cref{sec:meas-models}, we derive our radar and camera measurement models.
With the necessary preliminaries in place, we then define the full estimation problem in \Cref{sec:state}.

\subsection{Notation}
\label{sec:notation}

Latin and Greek letters (e.g., $a$ and $\alpha$) denote scalar variables, while boldface lower- and uppercase letters (e.g., $\Vector{x}$ and $\Matrix{\Theta}$) denote vectors and matrices, respectively.
A parenthesized superscript pair, for example, $\Matrix{A}^{(i, j)}$, indicates the $i$th row and the $j$th column of the matrix $\Matrix{A}$. 
A three-dimensional reference frame is designated by $\CoordinateFrame{}$.
The translation vector from point $a$ (often a reference frame origin) to $b$, expressed in $\CoordinateFrame{a}$, is denoted by $\!\Vector{r}_a^{ba}$.
The translational velocity vector of point $b$ relative to point $a$, expressed in $\CoordinateFrame{a}$, is denoted by $\Vector{v}_a^{ba}$. 
The angular velocity of frame $\CoordinateFrame{a}$ relative to a frame $\CoordinateFrame{i}$, expressed in $\CoordinateFrame{a}$, is denoted by $\!\Vector{\omega}_a^{ai}$.

We denote rotation matrices by $\Matrix{R}$.
For example, $\Matrix{R}_{ab} \in \mathrm{SO}(3)$ defines the rotation from $\CoordinateFrame{b}$ to $\CoordinateFrame{a}$. 
We reserve $\Matrix{T}$ for $\LieGroupSE3$ transformation matrices.
For example, $\Matrix{T}_{ab}$ is the 4 $\times$ 4 homogeneous matrix that defines the rigid-body transform from frame $\CoordinateFrame{b}$ to $\CoordinateFrame{a}$.
Our $\LieGroupSE3$ matrix entries will generally be functions of time; we denote the transform from frame $\CoordinateFrame{b}$ to $\CoordinateFrame{a}$ at time $t$ by
\begin{equation}
\Matrix{T}_{ab}(t) = 
\bbm \Matrix{R}_{ab}(t) & \Vector{r}_a^{ba}(t) \\ \Transpose{\Vector{0}} & 1 \ebm,
\end{equation}
where $\Matrix{R}_{ab}(t) \in \mathrm{SO}(3)$ and $\Vector{r}_a^{ba}(t) \in \Real^3$.
We use $\Identity_{n}$ to denote the $n$-by-$n$ identity matrix.

The unary operator $^\wedge$ acts on $\Vector{r} \in \Real^3$ to produce a skew-symmetric matrix such that $\Vector{r}^\wedge\Vector{s}$ is equivalent to the cross product $\Vector{r} \times \Vector{s}$.
The operators $\exp(\cdot)$ and $\log(\cdot)$ map from the Lie algebra $\mathfrak{so}(3)$ to the Lie group $\mathrm{SO}(3)$ and vice versa, respectively \cite{barfoot2017state}.

\subsection{Continuous-Time Trajectory Representation}
\label{sec:b-splines}

Temporal calibration is most easily formulated as a continuous-time problem, in part because the batch optimization procedure incrementally time-shifts the measurements from one sensor.
In turn, we require the ability to query the pose of the radar or the camera at arbitrary points in time.
To enable this, we parameterize the trajectory of the radar-camera pair using the B-spline representation from Sommer et al.\ \cite{Sommer_Efficient_2019}.
This representation is briefly reviewed below. We refer readers to Sommer et al.\ \cite{Sommer_Efficient_2019}, de Boor \cite{1978_Boor_Splines}, and Qin \cite{1998_Qin_Splines} for additional details.

A B-spline of order $k$ is a function of one continuous parameter (e.g., time) and a finite set of control points; for brevity, we restrict our example here to control points $\{\Vector{p}_0, \dots, \Vector{p}_N \mid \Vector{p}_i \in \Real^d\}$.
In a uniformly-spaced B-spline, each control point is assigned a time (or \emph{knot}) $t_i = t_0 + i \Delta t$, where $t_0$ marks the beginning of the spline and $\Delta t$ is the time between knots.
Evaluating a $k^\text{th}$ order B-spline at time $t$, where $t_i \leq t < t_{i+1}$, requires the set of $k$ control points over the knot sequence $t_i, \dots, t_{i+k-1}$.
As a result, the end point of a B-spline of length $N$ and order $k$ is at time $t_{N - k + 1}$.
 
The first step in computing the value of a $k^\text{th}$ order B-spline at time $t$ is to convert $t$ to the `normalized' time $u = \frac{t - t_i}{t_{i+1} - t_i}$.
Given $u$, the value of the $k^\text{th}$ order B-spline is defined as
\begin{equation} \label{eq:rd_spline}
\Vector{p}(u) = \bbm \Vector{p}_i & \Vector{d}_1^i & \dots & \Vector{d}_{k-1}^i \ebm\tilde{\Matrix{M}}_{k}\Vector{u},
\end{equation}
where $\Vector{u}^T = [1 \; u \; u^2 \;\dots\; u^{k-1}]$ and $\Vector{d}_j^i = \Vector{p}_{i+j} - \Vector{p}_{i+j-1}$.
The elements of the $k\times k$ mixing matrix, $\tilde{\Matrix{M}}_{k}$, are defined by, 
\begin{align} \label{eq:mix}
\tilde{\Matrix{M}}^{(a,n)}_{k} &= \sum_{s=a}^{k-1}m^{(s,n)}_{k}, \\
\begin{split}
m^{(s,n)}_{k} &= \frac{C^n_{k-1}}{(k-1)!} \sum_{l=s}^{k-1}(-1)^{l-s}C_k^{l-s}(k-1-l)^{k-1-n}\\
& a,s,n \in \{0, \dots, k-1 \},
\end{split}
\end{align}
where scalar $C^i_{j} = \frac{j!}{i!(j-i)!}$.
Substituting $\Vector{\lambda}(u) = \tilde{\Matrix{M}}_{k}\Vector{u}$ into \Cref{eq:rd_spline} results in 
\begin{equation} \label{eq:rd_spline_simp}
\Vector{p}(u) = \Vector{p}_i + \sum_{j=1}^{k-1}\Vector{\lambda}_j(u)\Vector{d}^i_{j}.
\end{equation}
\Cref{eq:rd_spline_simp} can describe the smooth translation of a rigid-body in continuous time (see \Cref{fig:spline} in \Cref{sec:real-experiments} for an example).

While our development above focuses on vector space splines, B-splines can also be defined on Lie groups, including the group $\mathrm{SO}(3)$ of rotations,
\begin{equation}
\Matrix{R}(u) = \Matrix{R}_i\prod_{j=1}^{k-1}\exp(\lambda_j(u)\Vector{\phi}_j^i),
\end{equation}
where $\Matrix{R}_i$ is a control point of the rotation spline and $\Vector{\phi}^i_j = \log(\Matrix{R}_{i+j-1}^T\Matrix{R}_{i+j})$.
We use two B-splines, one on $\mathrm{SO}(3)$ and one on $\Real^3$, as our complete continuous-time representation of the radar-camera trajectory.

\subsection{Sensor Measurement Models}
\label{sec:meas-models}

In order to perform spatiotemporal calibration, we require a measurement model for the radar unit.
Radars emit EM waves that reflect off of surfaces in the environment.
Due to the relatively long EM wavelength used by radars, the reflected ``location'' of a target can vary based on the relative orientation between the radar and target \cite{richards_principle_2010}.
Additionally, multipath reflections can occur when the wave bounces off of multiple surfaces before returning to the radar receiver, which can bias measurements of targets and introduce false detections \cite{richards_principle_2010}.

\begin{figure}[t]
	\centering
	\begin{tikzpicture}
	\node[anchor=south west,inner sep=0] (image) at (0,0) {\includegraphics[width=0.9\columnwidth]{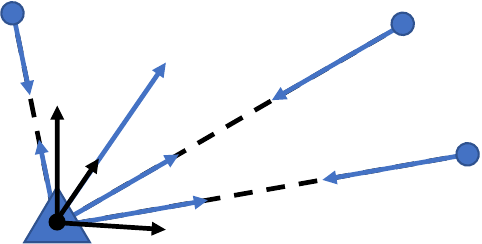}};
	\draw (0.0, 0.5) node {Radar};
	\draw (2.9, 0.25) node {$x$};
	\draw (1.5, 1.5) node {$y$};
	\draw (0.95, 2.5) node {$z$};
	\draw (3.0, 3.25) node {$\Vector{h}_r$};
	\draw (5.2, 1.4) node {$\dot{r}_1$};
	\draw (3.1, 1.1) node {$-\dot{r}_1$};
	\draw (4.3, 2.7) node {$\dot{r}_2$};
	\draw (2.5, 1.75) node {$-\dot{r}_2$};
	\draw (0.1, 2.7) node {$\dot{r}_3$};
	\draw (0.2, 1.2) node {$-\dot{r}_3$};
	\draw (7, 1.8) node {$(\theta_1, \phi_1)$};
	\draw (5.8, 3.85) node {$(\theta_2, \phi_2)$};
	\draw (1.1, 3.85) node {$(\theta_3, \phi_3)$};
	\end{tikzpicture}
	\vspace{1mm}
	\caption{Illustration of our radar measurement model. The radar EM wave reflects off of three (or more) non-collinear, stationary landmarks in the environment,  yielding azimuth, elevation, and range-rate measurements to each landmark. Using these data, we estimate the radar velocity relative to the world reference frame expressed in the radar reference frame.}
	\label{fig:ego-velocity}
	\vspace{-3mm}
\end{figure}
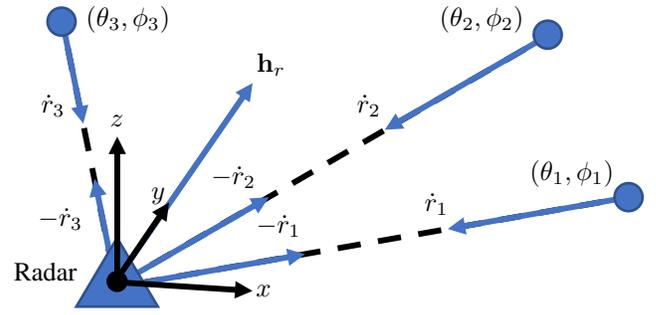

For each received reflection $l$ from an environmental feature (i.e., an object that we identify as a \emph{landmark}), the radar measures the range $r_l$, azimuth $\theta_l$, elevation $\phi_l$, and range-rate $\dot{r}_l$.
We assume that the observed landmarks are stationary with respect to a world frame, $\CoordinateFrame{w}$; measurements are resolved in the radar reference frame, $\CoordinateFrame{r}$. 
The range-rate measurement is the dot product between the velocity of the radar unit itself, $\Vector{h}_r \in \Real^3$, and the unit vector $\hat{\Vector{r}}_l \in \mathrm{S}^2$ defined by $\theta_l$ and $\phi_l$.
Given radar measurements to at least three non-collinear, stationary landmarks, the  unit direction vectors and their associated range-rates can be used to reconstruct the radar velocity, $\Vector{h}_r$, as shown in \Cref{fig:ego-velocity}. 

Stahoviak \cite{Stahoviak_Velocity_2019} and Doer et al.\ \cite{doer_reve_2020} demonstrate that, given $N > 3$ landmarks, one can estimate the ego-velocity of the radar by solving the over-constrained linear least-squares problem
\begin{equation} \label{eq:ego-problem}
\Vector{h}_r^\star = \min_{\Vector{h}_r} \; \Transpose{\Vector{e}_{ego}}\Vector{e}_{ego},
\end{equation}
where
\begin{equation} \label{eq:ego-meas}
\Vector{e}_{ego} = \Matrix{H}\Vector{x} - \Vector{y} = 
\bbm \Transpose{\hat{\Vector{r}}_0} \\ \vdots \\ \Transpose{\hat{\Vector{r}}_N} \ebm \Vector{h}_r -
\bbm \dot{r}_0 \\ \vdots \\\dot{r}_N \ebm.
\end{equation}
\Cref{eq:ego-meas} has its specific form because we wish to estimate the velocity of the radar with respect to the static world frame --- not vice versa. 
The estimated ego-velocity covariance is
\begin{equation} \label{eq:ego-cov}
\Vector{\Sigma}_{v} = \frac{(\Transpose{\Vector{e}_{ego}}\Vector{e}_{ego})(\Transpose{\Matrix{H}}\Matrix{H})}{N-3}.
\end{equation}

We use RANSAC \cite{Stahoviak_Velocity_2019,doer_reve_2020} and radar cross-section thresholding to remove outliers.
The two main sources of outliers are targets that move relative to the inertial reference frame and spurious multipath reflections.
Empirically, RANSAC successfully eliminates outliers from these two sources if the range-rates to the targets deviate significantly from other stationary landmarks that are close in terms of angle.
However, there are two subtle cases where multipath returns may appear to be valid measurements of stationary landmarks.
In the first case, the difference between the transmission and return angles is small.
In the second case, the transmission and return angles are symmetric about the radar ego-velocity direction (see Section 8.9 in \cite{richards_principle_2010}).
The returns from these reflections have a small radar cross-section and are rejected by cross-section thresholding.

Leveraging our continuous-time trajectory representation, the measurement model for the radar ego-velocity at time $t_j$ is
\begin{equation} \label{eq:radar-velocity-model}
\begin{split}
\Vector{h}_{r_j} & = - \dot{\Vector{r}}_r^{wr}(t_j + \tau) - \Vector{\omega}_r^{rw}(t_j + \tau)^{\wedge}\Vector{r}_r^{wr}(t_j + \tau) \\ & \qquad\qquad +\Vector{n}_{v_j}, \\
\Vector{n}_{v_j} & \sim  \NormalDistribution{\Vector{0}_{3\times 1}}{\Matrix{\Sigma}_{v_j}},
\end{split}
\end{equation}
where $\tau$ is the temporal offset of the radar measurements relative to the camera measurements and $\Vector{n}_{v_j}$ is the radar velocity measurement noise term. We assume that the noise is a zero-mean Gaussian with covariance matrix $\Matrix{\Sigma}_{v_j}$ (see \Cref{eq:ego-cov}).
From the radar ego-velocity model, the error residual is 
\begin{equation}\label{eq:radar-velocity-meas}
\begin{split}
\Vector{e}_{v_j} & = \Vector{h}_{r_j} + \dot{\Vector{r}}_r^{wr}(t_j + \tau) + \\ & \qquad\qquad \Vector{\omega}_r^{rw}(t_j + \tau)^{\wedge}\Vector{r}_r^{wr}(t_j + \tau) - \Vector{n}_{v_j}.
\end{split}
\end{equation}

To estimate the ego-motion of the camera, we use a monocular simultaneous localisation and mapping (SLAM) algorithm that operates independently of the radar. By observing fixed landmarks in the environment, monocular SLAM is capable of determining the transformation between the camera reference frame, $\CoordinateFrame{c}$, and the world frame, $\CoordinateFrame{w}$, up to an unknown scale factor $\alpha$\cite{chiuso2002_structure}.
Our (scaled) camera pose measurement model is given by
\begin{equation}
\begin{split}
\Matrix{R}_{cw,t_k} &= \exp(\Vector{n}_{r,k})\Matrix{R}_{cr}\Transpose{\Matrix{R}_{wr}(t_k)},\\
\Vector{n}_{r,k} &\sim \NormalDistribution{\Vector{0}_{3\times 1}}{\Matrix{\Sigma}_r},\\
\end{split}
\end{equation}
\begin{equation}
\begin{split}
\Vector{r}_{c,t_k}^{wc} &= \alpha(\Matrix{R}_{cr}\Vector{r}_r^{wr}(t_k) + \Vector{r}_c^{rc}) + \Vector{n}_{t,k},\\
\Vector{n}_{t,k} &\sim \NormalDistribution{\Vector{0}_{3\times 1}}{\Matrix{\Sigma}_t},
\end{split}
\end{equation}
where $\Vector{n}_{r,k}$ and $\Vector{n}_{t,k}$ are zero-mean Gaussian noise terms of covariances matrice $\Matrix{\Sigma}_r$ and $\Matrix{\Sigma}_t$ for the camera rotation and translation measurements.
The resulting error equations are
\begin{align}
\Vector{e}_{r,t_k} &= \log(\Matrix{R}_{cw,t_k}\Matrix{R}_{wr}(t_k)\Transpose{\Matrix{R}_{cr}}),\\
\Vector{e}_{t,t_k} &= \Vector{r}_{c,t_k}^{wc} - \alpha(\Matrix{R}_{cr}\Vector{r}_r^{wr}(t_k) + \Vector{r}_c^{rc}) - \Vector{n}_{t,k}.
\end{align}
 
We note that a monocular visual odometry (VO) algorithm (i.e., localization without loop closure) can provide camera ego-motion estimates, but visual drift will bias these estimates and decrease calibration accuracy.
Also, the measured radar ego-velocity is a local property of a trajectory and cannot fully correct for pose errors induced by visual drift.
  
\subsection{The Spatiotemporal Calibration Problem}
\label{sec:state}
The set of parameters, $\Vector{x}$, that we wish to estimate are the spline control points ($\Vector{r}_{0 \dots N} \in \Real^3$, $\Matrix{R}_{0 \dots N} \in \LieGroupSO{3}$), the extrinsic calibration parameters ($\Matrix{R}_{cr}, \Vector{r}_c^{rc}$), the camera translation scale factor ($\alpha$), and the temporal offset ($\tau$),
\begin{equation}
\begin{split}
\Vector{x} = & \left\{\begin{matrix} \Vector{r}_0, & \dots, & \Vector{r}_N, & \Matrix{R}_0, & \dots, & \Matrix{R}_N, \end{matrix}\right. \\
	& \; \left.\begin{matrix}\Matrix{R}_{cr}, & \Vector{r}_c^{rc}, & \alpha, & \tau \end{matrix}\right\}.
\end{split}
\end{equation}
Given $N_r$ radar measurements and $N_c$ camera measurements, we minimize the following cost function,
\begin{equation}
\begin{aligned}
\Vector{x}^\star = \min_{\Vector{x}} \; & \sum_{j=1}^{N_r} \Transpose{\Vector{e}_{v_j}}\Matrix{\Sigma}_{v_j}^{-1}\Vector{e}_{v_j} + \\ &\sum_{k=1}^{N_c} \Transpose{\Vector{e}_{r, t_k}}\Matrix{\Sigma}_r^{-1}\Vector{e}_{r, t_k} + \Transpose{\Vector{e}_{t, t_k}}\Matrix{\Sigma}_t^{-1}\Vector{e}_{t, t_k}.
\end{aligned}
\end{equation}
We perform this minimization using the Ceres solver, a standard nonlinear least squares solver \cite{ceres-solver}.
The ability to calibrate all of the relevant parameters depends upon the identifiability of problem, which we discuss in the next section.

\section{Identifiability}
\label{sec:ident}

In this section, we show that the calibration problem is identifiable given sufficient excitation of the radar-camera system.
Our approach is to determine the observability, or `instantaneous identifiability,' of the system at several different points in time, assuming that the system follows a varying trajectory.
We consider local identifiability (cf.\ locally weak observability) along a trajectory segment in \Cref{subsec:radar-cam}, after introducing the requisite observability rank condition in \Cref{subsec:obs-rank}.
A similar approach has been taken in \cite{2014_Li_Online} and \cite{2015_Hewitt_Towards} and elsewhere.
In \Cref{subsec:degenerate}, we describe several `degenerate' motions for which the identifiability condition does not hold.
We leave the complete characterization of the set of unidentifiable trajectories as future work.

\subsection{The Observability Rank Condition}
\label{subsec:obs-rank}

We make use of the criterion from Hermann and Krener \cite{hermann_nonlinear_1977} as part of our identifiability analysis.
A system $S$, written in control-affine form as 
\begin{equation}
S\,\begin{cases}
\dot{\Vector{x}} = \Vector{f}_0(\Vector{x}) + \sum_{j=1}^p \Vector{f}_j(\Vector{x})u_j \\
\Vector{y} = \Vector{h}(\Vector{x})
\end{cases},
\end{equation}
with the drift vector field $\Vector{f}_0(\Vector{x})$ and control inputs $u_j$ (for $j = 1, \dots, p$), is locally weakly observable if the matrix $\Matrix{O}$ of the gradients of the Lie derivatives with respect to the system state has full column rank.

The Lie derivative, or directional derivative, of a smooth scalar function $h$ with respect to the smooth vector field $\Vector{f}$ at the point $\Vector{x}$ is
\begin{equation}
L_{\Vector{f}} h(\Vector{x}) = \nabla_{\Vector{f}}h(\Vector{x}) = \frac{\partial h(\Vector{x})}{\partial \Vector{x}}\Vector{f}(\Vector{x}).
\end{equation}
The $n^\text{th}$ Lie derivative of $h$ with respect to $\Vector{x}$ along $\Vector{f}$ is defined recursively as
\begin{equation}
L_{\Vector{f}}^n h(\Vector{x}) = \frac{\partial L_{\Vector{f}}^{n-1} h(\Vector{x})}{\partial \Vector{x}}\Vector{f}(\Vector{x}),
\end{equation}  
where $L^0 h(\Vector{x})=h(\Vector{x})$.
We note that the matrix $\Matrix{O}$ has an infinite number of rows, but it is sufficient to show that a finite subset of rows yield a matrix of full column rank.

\subsection{Identifiability of Radar-Camera Calibration}
\label{subsec:radar-cam}

We begin by simplifying the state (and parameter) vector that we aim to estimate.
We are able to measure the camera pose up to scale \cite{chiuso2002_structure} and the radar velocity in the radar frame \cite{Stahoviak_Velocity_2019}. 
Since we are working in continuous time (or, roughly equivalently, if there are a sufficient number of closely-spaced radar and camera measurements), then the scaled velocity of the camera in the camera reference frame $\alpha\Vector{v}_c^{cw}(t_i)$, the angular velocity of the camera $\Vector{\omega}_c(t_i)$ in the camera frame, the radar velocity $\Vector{v}_r^{rw}(t_i + \tau)$ in the radar frame, and the time derivative of the radar velocity $\dot{\Vector{v}}(t_i + \tau)$ in the radar frame are all available.
For the purposes of identifiability, we are able to define the following, modified measurement model,
\begin{equation}\label{eqn:obs_meas_fun}
\Vector{h}(t_i) = \alpha (\Matrix{R}_{cr}\Vector{v}_r^{rw}(t_i + \tau) - \Vector{\omega}_c(t_i)^{\wedge}\Vector{r}_c^{rc}),
\end{equation}
where $\Vector{h}(t_i)$ is the scaled linear velocity of the camera ($\Vector{v}_c^{cw}$) and $\Vector{\omega}_{c}$ is the angular velocity of the camera, both relative to the camera frame.
This modified measurement model does not directly rely on the pose of the radar, thus simplifying the set of parameters that we wish to determine to
\begin{equation}
\tilde{\Vector{x}} = \{\begin{matrix} \Vector{r}_c^{rc}, & \Matrix{R}_{cr}, & \alpha, & \tau \end{matrix}\}.
\end{equation}
 
To decrease the notational burden, we drop the superscripts and subscripts defining the velocities and extrinsic transform parameters.
The gradient of the zeroth-order Lie derivative of the $i$\hspace{0.05em}th measurement is
\begin{equation} \label{eqn:obs_meas}
\begin{aligned}
\nabla_{\tilde{\Vector{x}}}L_0\Vector{h}(t_i) = & \left[\begin{matrix} -\alpha\Vector{\omega}(t_i)^\wedge & -\alpha(\Matrix{R}\Vector{v}(t_i + \tau))^\wedge\Matrix{J} \end{matrix}\right.\\
 & \left.\begin{matrix} \Matrix{R}\Vector{v}(t_i + \tau) - \Vector{\omega}(t_i)^{\wedge}\Vector{r} & \alpha \Matrix{R}\dot{\Vector{v}}(t_i + \tau) \end{matrix}\right],
\end{aligned}
\end{equation}
where $\Matrix{J}$ is the Lie algebra left Jacobian of $\Matrix{R}_{cr}$ \cite{barfoot2017state}.
Since the parameters of interest are constant with respect to time, we are able to stack the gradients of several Lie derivatives (at different points in time) to form the observability matrix,
\begin{equation} \label{eq:obs_mat}
\Matrix{O} = \bbm \nabla_{\tilde{\Vector{x}}}L_0\Vector{h}(t_1) \\ \nabla_{\tilde{\Vector{x}}}L_0\Vector{h}(t_2) \\ \nabla_{\tilde{\Vector{x}}}L_0\Vector{h}(t_3) \ebm,
\end{equation}
which, using block Gaussian elimination, can be shown to have full column rank when three or more sets of measurements are available.\footnote{We omit the full derivation for brevity, and note that the rank condition can be verified in this case using any symbolic algebra package.}

Two comments regarding the analysis are in order. 
First, we note that the analysis is simplified by considering the modified measurement equation only, which avoids the use of higher-order Lie derivatives.
Second, there is a subtlety involved in stacking the gradients of the Lie derivatives at different points in time. 
The modified measurement equation depends upon the time derivatives of the camera pose and the radar ego-velocity---this implies that, although we do not consider specific control inputs, the system dynamics must be non-null. 
Stated differently, varied motion of the radar-camera pair is necessary to ensure identifiability; we discuss this requirement further in the next section. Further, it is worth noting that the the observation times must span the temporal offset period \cite{2021_Kelly_Question}.

\subsection{Degenerate Motions}
\label{subsec:degenerate}

There are motions that cause the matrix $\Matrix{O}$ in \Cref{eq:obs_mat} to lose full column rank.
First, the Lie derivatives include linear and rotational velocities and accelerations, so the matrix will lose full column rank when the system is stationary with respect to the world frame or moving with constant linear or angular velocity.
Second, in Wise et al.\ \cite{wise_continuous-time_2021}, we showed that the system must undergo rotation about two nonparallel axes in order for the observability matrix to be full rank. 
This requirement also applies to the present analysis. To show this, we can align the angular velocity and angular acceleration vectors by substituting $\Vector{\alpha}_c^{ci} = \eta\,\Vector{\omega}_c^{cw}$, where $\eta$ is an arbitrary constant, into \Cref{eq:obs-matrix}.
This substitution is equivalent to asserting that the system rotates about one axis only, resulting in an observability matrix that is rank-deficient.

\section{Simulation Studies}
\label{sec:sim-experiments}

\begin{figure}[t] 
	\centering
	\includegraphics[width=0.9\columnwidth]{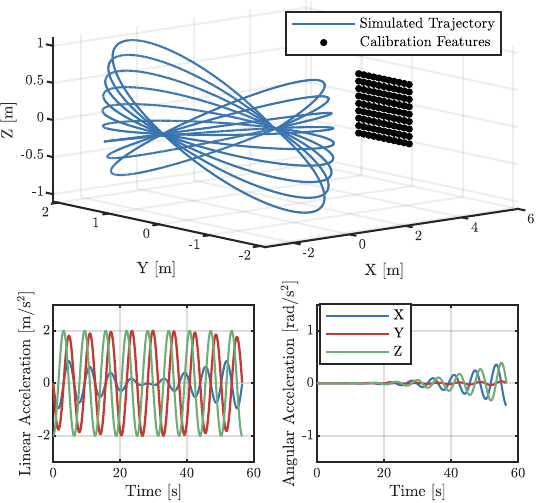}
	\caption{High linear and low angular velocity trajectory (top) for the simulation experiments, with associated linear (bottom left) and angular acceleration (lower right) plots.}
	\label{fig:sim-trajectory}
\end{figure}

\begin{figure}[t] 
	\centering
	\includegraphics[width=0.9\columnwidth]{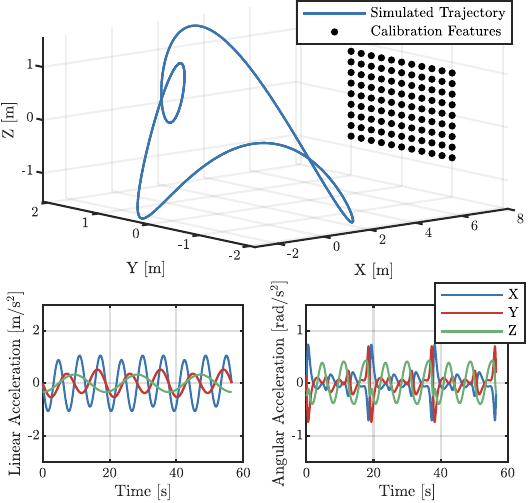}
	\caption{Low linear and high angular velocity trajectory (top) for the simulation experiments, with associated linear (bottom left) and angular acceleration (lower right) plots.}
	\label{fig:sim-trajectory-new}
\end{figure}

In order to test the robustness of our algorithm to measurement noise, we carried out a series of simulation studies. 
We generated two simulated camera-radar datasets using two different trajectories and with varying amounts of (synthetic) noise (see Figures \ref{fig:sim-trajectory} and \ref{fig:sim-trajectory-new}). 
The nominal (noise-free) trajectories were selected to ensure sufficient excitation of the camera-radar pair.
The median linear and rotational velocities for the trajectory shown in \Cref{fig:sim-trajectory} were, respectively, higher and lower than the velocities for the trajectory shown in \Cref{fig:sim-trajectory-new}.

After constructing the trajectories, we computed the simulated radar ego-velocity and camera pose measurements.
Since radar measurements are antenna configuration- and environment-specific, these measurements were not generated at the EM propagation level.
Radar ego-velocity measurements (i.e., $\Vector{h}_{r_k}$) were computed using the known linear and rotational velocities defined by the trajectory.
Consequently, our simulated radar measurements generalize to any radar and environment that produce an unbiased 3D ego-velocity estimate.
Simulated camera pose measurements (i.e., $\Matrix{R}_{cw,t_k}$ and $\Vector{t}_{c, t_k}^{wc}$) were derived from simulated observations of a series of landmark points, arranged in a 2D grid.
This configuration of points matches the configuration of a standard `checkerboard' camera calibration target.
In the targetless setting, we can only estimate the position of the camera up to an unknown scale \cite{chiuso2002_structure}, so the checkerboard tracking algorithm is given an incorrect size for the checkerboard squares.

For each individual simulation, we added zero-mean Gaussian noise to the radar ego-velocity measurements ($\Matrix{\Sigma}_{v_j} = \sigma_r^2\,\Identity_{3 \times 3}$) and to the camera measurements of the checkerboard corners on the simulated image plane ($\Matrix{\Sigma}_{p_k} = \sigma_c^2\,\Identity_{2 \times 2}$). 
In our experiments, we adjusted the radar ego-velocity noise standard deviation ($\sigma_r$) between 0.05 m/s and 0.15 m/s.
Based on our real-world experiments (see \Cref{sec:real-experiments}), we have found that the radar ego-velocity measurement noise is closer to the lower end of this range, unless the environment is sparse and too few valid radar returns are captured.
We adjusted the standard deviation of the noise added to the measured checkerboard corner coordinates ($\sigma_c$) between 0.2 and 0.4 pixels; these noise levels are similar to the observed noise in our real-world experiments \cite{wise_continuous-time_2021}. 

\begin{figure*}[t]
	\centering 
	\includegraphics[width=\textwidth]{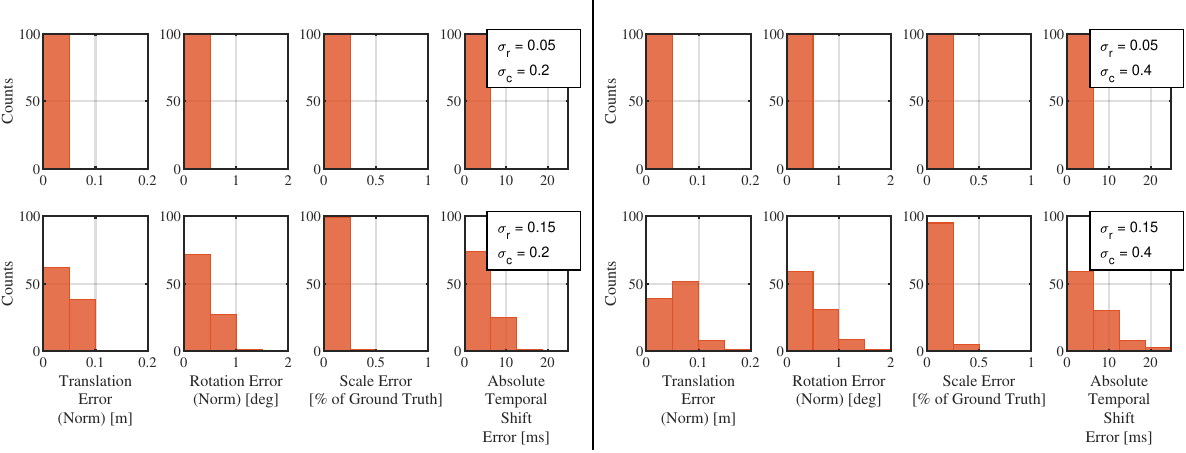}
	\caption{High linear and low angular velocity trajectory calibration results from our simulation experiments. Each subplot is a histogram of the error between the estimated and true parameter values for 100 trials at a given level of measurement noise. Each row presents the results for a level of measurement noise. The levels of noise are a combination of two radar measurement noise levels ($\sigma_r = 0.05 \text{ or } 0.15$ m/s) and two camera pixel measurement noise levels ($\sigma_c = 0.2 \text{ or } 0.4$ pixels).}
	\label{fig:sim-results}
\end{figure*}

\begin{figure*}[t]
	\centering 
	\includegraphics[width=\textwidth]{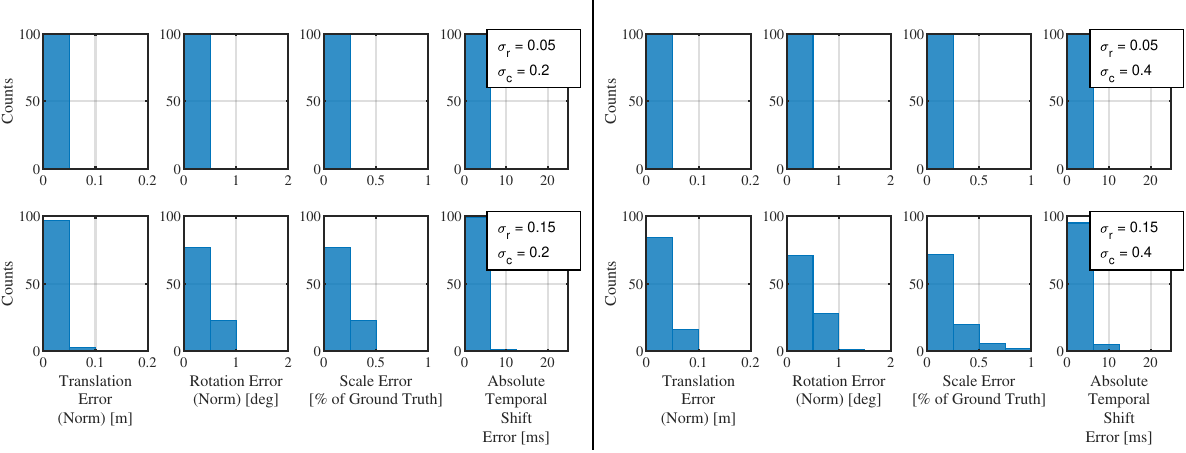}
	\caption{Low linear and high angular velocity trajectory calibration results from our simulation experiments. Each subplot is a histogram of the error between the estimated and true parameter values for 100 trials at a given level of measurement noise. Each row presents the results for a level of measurement noise. The levels of measurement noise are a combination of two radar measurement noise levels ($\sigma_r = 0.05 \text{ or } 0.15$ m/s) and two camera pixel measurement noise levels ($\sigma_c = 0.2 \text{ or } 0.4$ pixels).}
	\label{fig:sim-results-new}
\end{figure*}

The error distributions for the spatial calibration parameter estimates ($\Matrix{R}_{cr}, \Vector{r}_c^{rc}$), scale factor ($\alpha$), and temporal offset ($\tau$) are shown in Figures \ref{fig:sim-results} and \ref{fig:sim-results-new}, across  100 simulation trials for each (nominal) trajectory.
For the high linear and low angular velocity trajectory, even in the high-noise regime, the error in the rotation and scale estimates remains less than two degrees and one percent, respectively.
However, high levels of noise in the radar ego-velocity measurements result in substantially larger (and more widely distributed) errors in the estimate of the relative translation of the sensors and of the temporal offset; the errors can be as large as 15 cm and 30 ms, respectively.
This sensitivity indicates that, prior to use in our algorithm, the radar data should be filtered to remove high-noise measurements whenever possible. 

If the system follows the low linear and high angular velocity trajectory in \Cref{fig:sim-trajectory-new}, then radar data filtering may not be necessary.
As shown in \Cref{fig:sim-results-new}, calibrating the radar along this trajectory results in similar scale and rotation estimation accuracy as for the other trajectory, but drastically improves the translation and temporal offset estimates; the errors are within 10 cm and 10 ms, respectively.
Additionally, our algorithm achieves a comparable spatial calibration accuracy to Doer et al.\ \cite{doer_radar_2020} on the noisier radar ego-velocity data.
However, the high angular velocity trajectory is challenging for real-world camera localization and the amount of excitation is not necessary if the radar data are sufficiently accurate.

\section{Real-World Experiments}
\label{sec:real-experiments}
 
To verify the performance and accuracy of our algorithm, we carried out a series of real-world experiments involving three different radar-camera systems.
We discuss the various systems and their implementation details in \Cref{sec:exp-info}.
In \Cref{sec:our-rig-experiments}, we show that the set of spatiotemporal calibration parameters estimated by our algorithm have a similar level of alignment accuracy as the parameters estimated by the target-based method of Per\v{s}i\'{c} et al.\ \cite{persic_spatiotemporal_2021}.
In \Cref{sec:IRS-experiments}, we demonstrate how spatiotemporal calibration can improve the performance of camera-radar-IMU odometry. 
Finally, in \Cref{sec:vehicle-experiments}, we evaluate the accuracy of our algorithm in a challenging situation involving sensors mounted on an autonomous vehicle.

\subsection{Data Collection and Data Preprocessing}
\label{sec:exp-info}

The data collection systems are different for each experiment, but each system includes at least one radar and one camera.
The system discussed in \Cref{sec:our-rig-experiments} is a handheld rig that incorporates a Texas Instruments (TI) AWR1843BOOST radar and Point Grey Flea3 USB camera. The measurement update rates for the sensors are 20 Hz and 30 Hz, respectively.
For the experiments in \Cref{sec:IRS-experiments}, the data are from the publicly available IRS Radar Thermal Visual Inertial dataset \cite{doer_rrxio_2021}.
The data collection system \cite{doer_rrxio_2021} is a handheld rig that mounts on a drone, where measurements are acquired from a TI IWR6843AOP radar, an IDS UI-3241 camera, and an Analog Devices ADIS16448 IMU, operating at frequencies of 10 Hz, 20 Hz, and 409 Hz, respectively.
Doer et al.\ \cite{doer_rrxio_2021} provides additional details about this system.
In \Cref{sec:vehicle-experiments}, the data collection system \cite{2023_burnett_boreas} includes a vehicle-mounted TI AWR1843BOOST radar and three Point Grey Flea3 GigE cameras operating at frequencies of 25 Hz and 16 Hz, respectively. 

In our real-world experiments, we use two similar radars that primarily differ in their angular resolutions.
If two targets have identical ranges and range-rates but are separated by less than the angular resolution, then the targets will blend together, biasing the radar measurement output.
The AWR1843BOOST has azimuth and elevation resolutions of 15$^\circ$ and 58$^\circ$, respectively, while the IWR6843AOP has azimuth and elevation resolutions of 30$^\circ$.
As we show in Sections \ref{sec:our-rig-experiments}, \ref{sec:IRS-experiments}, and \ref{sec:vehicle-experiments}, our algorithm is capable of calibrating both radars even though they have differing angular resolutions.
For additional information on the radars used in our experiments, we refer the reader to the AWR1843BOOST and IWR6843AOP user manuals \cite{AWR1843BOOST, IWR6843AOP}.

To ensure reliable ego-velocity estimation in our experiments, we set the maximum measurable range-rate and constant false alarm rate (CFAR) thresholds for our radar units.
The maximum range-rate of the radar must be set above the maximum velocity of the data collection platform because the ego-velocity estimates will saturate at this value.
However, an inverse relationship exists between the maximum range-rate and maximum range settings and these must be properly balanced for the operating environment \cite{richards_principle_2010}.
The on-board radar preprocessing pipeline incorporates a CFAR detector that differentiates targets from background noise in the received EM signal \cite{richards_principle_2010}.
Since the definition of background noise is also environment-dependent, we set the CFAR threshold to ensure that the ego-velocity estimator returned a sufficient number of inliers while minimizing the number of outliers.
Before each experiment, we performed a series of `test' data collection runs to tune these settings, making sure that the ego-velocity estimates were not saturating, that there were at least 15 inliers for each measurement, and that the inlier-to-outlier ratio was above 50\%.

There are three data preprocessing steps for the experiments discussed in Sections \ref{sec:our-rig-experiments} and \ref{sec:IRS-experiments}, while the experiment in \Cref{sec:vehicle-experiments} requires a fourth preprocessing step.
Prior to estimating the calibration parameters using our algorithm, we first determine radar ego-velocity estimates using the algorithm from \cite{doer_reve_2020}.\footnote{Available at: https://github.com/christopherdoer/reve}
Second, we rectify the camera images to remove lens distortion effects.
Third, we use the feature-based, monocular SLAM algorithm ORB-SLAM3 \cite{campos_orb3_2021} to provide an initial estimate of the (arbitrarily-scaled) pose of the camera at the time of each image acquisition.
While camera pose estimation is possible with any monocular SLAM, we chose this package for its robustness and accuracy \cite{campos_orb3_2021}. 
Finally, for the experiment in \Cref{sec:vehicle-experiments}, we remove outlier radar ego-velocity and camera pose estimates using a median filter.
The median filter computes the local median and standard deviation of the signals across a window of time---200 ms and 850 ms for the radar and the camera, respectively.
If the measurement at the centre of the window is greater than a chosen threshold from the median, the measurement is treated an outlier.
For the tests in \Cref{sec:vehicle-experiments}, the threshold is set to three standard deviations from the median, since this value eliminates gross outliers without removing noisy, but valid, portions of the signals.
We found that this step was necessary to ensure data integrity.

\subsection{Handheld Rig Experiment}
\label{sec:our-rig-experiments}

In this experiment, we compared the calibration parameters estimated by our algorithm against the parameters determined by the target-based method in Per\v{s}i\'{c} et al.\ \cite{persic_spatiotemporal_2021}.
To compare the two approaches, we used a handheld rig to collect a dataset consisting of two parts: one part with no visible calibration targets (for our algorithm) and one part with visible targets for target-based calibration.
We collected both parts during one continuous run, without power-cycling the sensors.
Our quality metric in this case is based on the results from target-based calibration (which can be treated as the `gold standard,' effectively).

\begin{figure}[t] 
	\centering
	\begin{tikzpicture}
	\node[anchor=south west,inner sep=0] (image) at (0,0) {\includegraphics[width=\columnwidth]{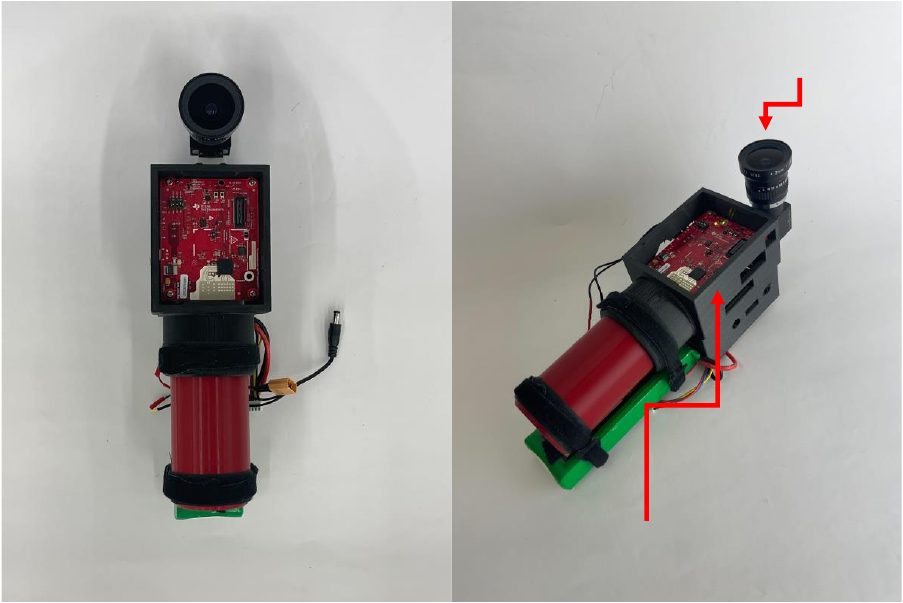}};
	\node (Radar) [rectangle,draw,fill=white, align=center] at (6.3, 0.48) {AWR1843BOOST\\3D mm-Wave Radar};
	\node (Camera) [rectangle,draw,fill=white, align=center] at (7.67, 5.43) {Point Grey\\Flea3 Cameras};
	\end{tikzpicture}
	\caption{Two pictures of our handheld sensor rig. The left image is a front view and the right image is an isometric view of the radar-camera unit. The radar antennas are mounted in the white area on the red circuit board. From our CAD model of the handheld rig, the radar-camera translation parameters (i.e., the components of $\Vector{r}_c^{rc}$) are $r_x = 0.1$, $r_y = 10.5$, and $r_z = -1.0$ cm.}
	\label{fig:handrig}
\end{figure}

\begin{figure}[t] 
	\centering
	\includegraphics[width=\columnwidth]{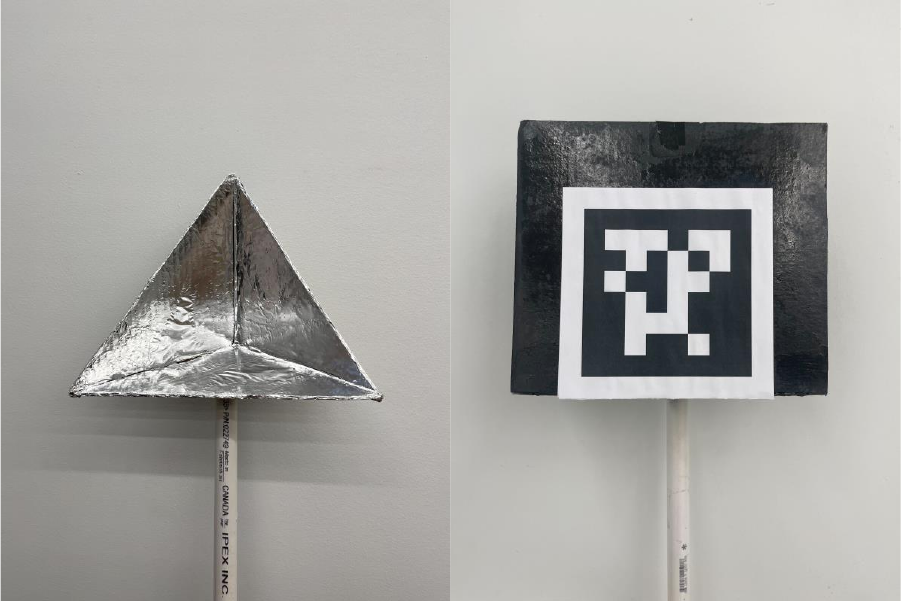}
	\caption{Our specialized retroreflective radar target used for calibration verification. The left image shows the retroreflector alone, while the right image shows an AprilTag mounted to a flat cardboard backing that is attached to the front of the retroreflector. The cardboard material is fully transparent to the radar EM wave.}
	\label{fig:target}
\end{figure}

\begin{figure}[h] 
	\centering
	\includegraphics[width=\columnwidth]{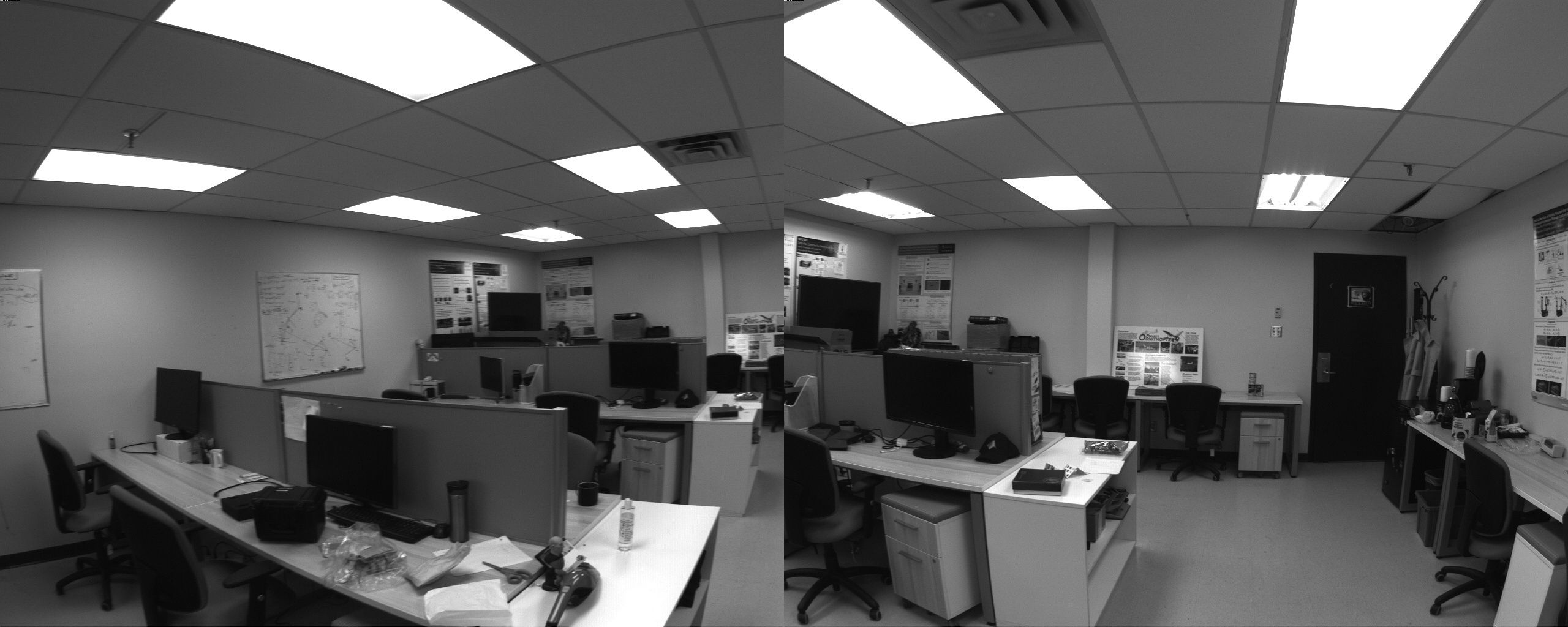}
	\caption{Images from our handheld sensor rig calibration dataset, showing two views of the feature-rich indoor test environment.}
	\label{fig:office_env}
\end{figure}

\begin{figure}[t] 
	\centering
	\includegraphics[width=\columnwidth]{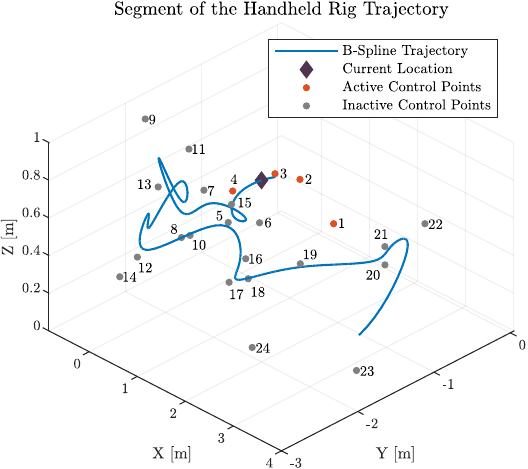}
	\caption{A segment of the estimated $\Vector{r}_r^{wr}$ B-spline for the handheld rig during calibration. The purple diamond is position of the rig 6.3 s from the start of the trajectory. The active control points at 6.3 s are shown in orange. As the rig continues along the trajectory, the active control points change.}
	\label{fig:spline}
	\vspace{-3mm}
\end{figure}

We used the first part of the dataset to perform targetless radar-camera calibration with our algorithm.
The procedure consisted of moving the sensor rig, shown in \Cref{fig:handrig}, throughout the office environment shown in \Cref{fig:office_env}.
A segment of the trajectory recovered by our algorithm is plotted in \Cref{fig:spline}.
Then, we used the second part of the dataset to perform target-based calibration with the algorithm described in Per\v{s}i\'{c} et al.\ \cite{persic_spatiotemporal_2021}.
In this case, the procedure consisted of moving a trihedral retroreflective target, shown in \Cref{fig:target}, in front of the stationary radar-camera rig.
The second part of the dataset was also used to evaluate the relative accuracy of the parameters estimated by both algorithms.

Our trihedral retroreflective target is specially constructed for calibration evaluation, consisting of a trihedral radar retroreflective `corner' and a visual AprilTag \cite{olson_tags_2011} pattern printed on paper  (which is EM-transparent).
The target, shown in \Cref{fig:target}, has the AprilTag mounted in front of the retroreflector.
Using the known AprilTag scale, the pose of the camera relative to the AprilTag reference frame can be established.
The distance from the origin of the AprilTag frame to the corner of the retroreflector is also known.
During data collection, we kept the reflector opening pointed at the radar to ensure a consistent radar reflection.  

We quantify the calibration accuracy based on a `reprojection error' metric.
The reprojection error is the distance between the position of the retroreflector corner predicted from the camera observations and the position measured by the radar, both expressed in the radar frame.
The retroreflector is more consistently detected than the AprilTag, and so we linearly interpolate the measured position of the trihedral retroreflector corner to the image timestamps.

\begin{figure}[b] 
	\centering
	\includegraphics[width=\columnwidth]{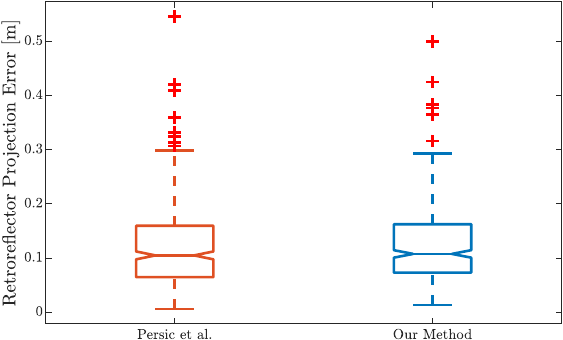}
	\caption{The reprojection error distributions for the state-of-the-art method in \cite{persic_spatiotemporal_2021} and ours.}
	\label{fig:compare-results}
\end{figure}

Overall, our algorithm achieves results that are comparable to the method from Per\v{s}i\'{c} et al.\ \cite{persic_spatiotemporal_2021}. 
\Cref{tab:handheld-params} shows that the estimated translation and rotation are, per axis, within 1.6 cm and 3 degrees, respectively, of the values estimated by the target-based method.
Additionally, our estimated temporal offset differs from the target-based method by only 6 ms.
\Cref{fig:compare-results} shows that our algorithm, in a completely targetless manner, produces a reprojection error distribution with a median that is only 3 mm larger than the target-based method.

\begin{table}[b]
	\scriptsize
	\centering
	\caption{Calibration parameters for our handheld dataset. The values in each row are estimated by a different algorithm. The rotation between the sensors is given in roll-pitch-yaw (i.e., $\theta_x$, $\theta_y$, $\theta_z$) Euler angle form.}
	\label{tab:handheld-params}
	\begin{threeparttable}
		\begin{tabular}{l@{\hspace{0.8\tabcolsep}}c@{\hspace{0.8\tabcolsep}}c@{\hspace{0.8\tabcolsep}}c@{\hspace{0.8\tabcolsep}}c@{\hspace{0.8\tabcolsep}}c@{\hspace{0.8\tabcolsep}}c@{\hspace{0.8\tabcolsep}}c@{\hspace{0.8\tabcolsep}}}
			\toprule
			& $r_x$ [cm] & $r_y$ [cm] & $r_z$ [cm] & $\theta_x$ [rads] & $\theta_y$ [rads] & $\theta_z$ [rads] & $\tau$ [ms]\\
			\midrule
			Per\v{s}i\'{c} \cite{persic_spatiotemporal_2021} & -1.60 & 11.9 & -5.02 & -1.59 & 0.07 & -3.12 & -63.8  \\
			Ours & -0.48 & 12.2 & -3.42 & -1.62 & 0.02 & -3.15 & -57.9 \\
			\bottomrule
		\end{tabular}
	\end{threeparttable}
\end{table}

\subsection{IRS Radar Thermal Visual Inertial Datasets}
\label{sec:IRS-experiments}

In this section, we demonstrate the versatility of our approach by making use of our estimated calibration parameters to improve the accuracy of camera-radar-IMU odometry.
Specifically, we evaluate on the dataset and against the camera-radar-IMU odometry algorithm, known as RRxIO, described by Doer and Trommer in \cite{doer_rrxio_2021}.
The extrinsic calibration parameters that accompany the IRS dataset were determined using the radar-IMU extrinsic calibration process described in \cite{doer_radar_2020} with ad hoc temporal calibration.
Post hoc calibration of the radar and camera is challenging because the test environments do not contain any trihedral reflectors and the motion of the sensor platform is constrained (i.e., there are no deliberate excitations for calibration).
To the best of the authors' knowledge, our approach is the only technique that can estimate all of the spatiotemporal calibration parameters for the dataset described in \cite{doer_rrxio_2021}.

We chose to calibrate (and to evaluate calibration quality) for three of nine trajectories in the IRS dataset: \texttt{Gym}, \texttt{MoCap Easy}, and \texttt{MoCap Medium}. 
Data were collected in two environments with varying numbers of features: a large, sparse gymnasium and a feature-rich office setting.
For the other six trajectories, poor lighting conditions and rapid motions caused ORB-SLAM3 to fail.
To evaluate on a given trajectory, we compute the radar-camera spatiotemporal calibration parameters using our algorithm, and then run RRxIO on the same dataset with our estimated parameters. 
During evaluation, we disable the live `camera-to-IMU' extrinsic calibration algorithm that operates as part of RRxIO.
Using the known ground truth and the estimated RRxIO trajectories, we are able to determine the quality of our calibration using the following odometric error metrics: the relative translational root mean square error (RMSE RTE), relative rotational RMSE (RRE), absolute translational RMSE (ATE), and absolute rotational RMSE (ARE).

While the parameters estimated by our algorithm, listed in \Cref{tab:IROS_params},  are relatively close to the parameters provided in Doer and Trommer \cite{doer_rrxio_2021}, use of our parameters yields more accurate odometry estimates.
The estimated temporal offset for the \texttt{Gym} trajectory is the only large deviation from the values in Doer and Trommer \cite{doer_rrxio_2021}.
\begin{table}[t]
	\scriptsize 
	\centering
	\caption{Radar-IMU calibration parameters determined for three IRS trajectories and by RRxIO. The radar-IMU calibration parameters listed for our algorithm are a combination of the IRS IMU-camera parameters and our camera-radar parameters. The rotation between the sensors is given by roll-pitch-yaw (i.e., $\theta_x$, $\theta_y$, $\theta_z$) Euler angles.}
	\label{tab:IROS_params}
	\begin{threeparttable}
		\begin{tabular}{l@{\hspace{0.8\tabcolsep}}c@{\hspace{0.8\tabcolsep}}c@{\hspace{0.8\tabcolsep}}c@{\hspace{0.8\tabcolsep}}c@{\hspace{0.8\tabcolsep}}c@{\hspace{0.8\tabcolsep}}c@{\hspace{0.8\tabcolsep}}c@{\hspace{0.8\tabcolsep}}}
			\toprule
			& $r_x$ [cm] & $r_y$ [cm] & $r_z$ [cm] & $\theta_x$ [rads] & $\theta_y$ [rads] & $\theta_z$ [rads] & $\tau$ [ms]\\
			\midrule
			RRxIO & 6.00 & 4.00 & -4.00 & -3.14 & 0.02 & -1.59 & 8.00 \\
			ME$^\dag$ (ours) & 4.08 & 4.71 & -5.05 & -3.12 & 0.01 & -1.59 & 13.1 \\ 
			MM$^\dag$ (ours) & 3.90 & 4.46 & -5.63 & -3.11 & 0.01 & -1.59 & 15.4 \\
			Gym (ours) & 3.27 & 4.48 & -3.62 & -3.15 & -0.06 & -1.60 & 40.7\\
			\bottomrule
		\end{tabular}
		\begin{tablenotes}
		\item[$\dag$] These datasets are MoCap Easy (ME) and MoCap Medium (MM).
		\end{tablenotes}
	\end{threeparttable}	
\end{table}
\Cref{tab:IROS_results} reports the absolute and relative translation and rotation errors for the RRxIO trajectories after a yaw alignment.
The parameters estimated by our algorithm improve the translation error on all datasets and rotation error for two of the datasets.
Notably, the \texttt{Gym} dataset, which has the largest temporal offset, improves the most.

\begin{table}[b]
	\scriptsize
	\centering
	\caption{Odometry performance evaluation for RRxIO and for our algorithm on three IRS trajectories.}
	\label{tab:IROS_results}
	\begin{threeparttable}
		\begin{tabular}{lc@{\hspace{0.8\tabcolsep}}cc@{\hspace{0.8\tabcolsep}}cc@{\hspace{0.8\tabcolsep}}cc@{\hspace{0.8\tabcolsep}}c}
			\toprule
			& \multicolumn{2}{c}{\textbf{RTE [\%]}} & \multicolumn{2}{c}{\textbf{RRE [deg/m]}}  & \multicolumn{2}{c}{\textbf{ATE [m]}} & \multicolumn{2}{c}{\textbf{ARE [deg]}} \\ \cmidrule(lr){2-3} \cmidrule(lr){4-5} \cmidrule(lr){6-7} \cmidrule(lr){8-9}
			\textbf{Dataset} & RRxIO & Ours & RRxIO & Ours & RRxIO & Ours & RRxIO & Ours\\ 
			\midrule
			ME$^\dag$ & 0.809 & 0.669 & 0.084 & 0.089 & 0.177 & 0.144 & 1.567 & 1.918 \\
			MM$^\dag$ & 1.377 & 1.097 & 0.122 & 0.095 & 0.351 & 0.260 & 2.522 & 2.027 \\
			Gym &  1.170 & 0.752 & 0.076 & 0.054 & 0.308 & 0.195 & 2.087 & 1.349\\
			\bottomrule
		\end{tabular}
		\begin{tablenotes}
			\item[$\dag$] These datasets are MoCap easy (ME) and MoCap Medium (MM).
		\end{tablenotes}
	\end{threeparttable}
\end{table}

\subsection{Vehicle Experiments}
\label{sec:vehicle-experiments}

\begin{figure}[b] 
	\centering
	\begin{tikzpicture}
	\node[anchor=south west,inner sep=0] (image) at (0,0) {\includegraphics[width=0.70\columnwidth]{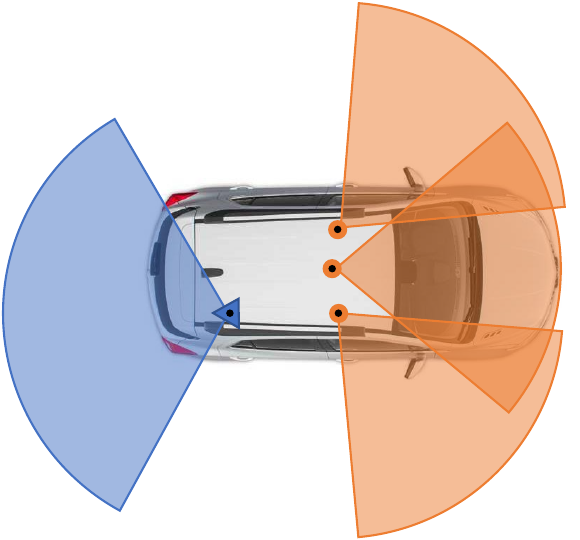}};
	\node (Radar) [rectangle,draw,fill=white, align=center] at (2.05, 1.15) {Radar\\Field of View};
	\node (Camera) [rectangle,draw,fill=white, align=center] at (3.15, 4.75) {Camera\\Fields of View};
	\end{tikzpicture}
	\caption{Fields of view of the radar and cameras sensors used for our vehicle experiments.}
	\label{fig:boreas_fov}
\end{figure}

In this section, we verify the accuracy of our calibration algorithm by estimating the distance between cameras mounted on an autonomous vehicle.
This task was challenging because, as shown in \Cref{fig:boreas_fov}, the radar-camera pairs do not share overlapping fields of view, so it is impossible to perform calibration using a target-based method.
Additionally, the constrained motion of the car results in a poorly conditioned problem (i.e., the minimum eigenvalue of the identifiability matrix in \Cref{eq:obs_mat} is close to zero).
The poor conditioning of the problem makes the estimated parameters very sensitive to sensor measurement noise, which can lead to inaccurate results.
To overcome the poor conditioning of this system, we add an extrinsic calibration prior,
\begin{equation}
\begin{aligned}
\Vector{e}_{prior} =& \log(\Matrix{T}_{cr}^{-1}\Matrix{T}_{cr,prior}), \\[2mm]
J_{prior} =& \Transpose{\Vector{e}_{prior}}\Matrix{\Sigma}_{prior}^{-1}\Vector{e}_{prior}, \\[2mm]
\Matrix{\Sigma}_{prior} =& \bbm \sigma_t^2\,\Identity_{3 \times 3} & \Matrix{0}_{3 \times 3} \\ \Matrix{0}_{3 \times 3} &  \sigma_\theta^2\,\Identity_{3 \times 3}\ebm,
\end{aligned}
\end{equation}
to the optimization problem.
For our experiments, the prior for the extrinsic calibration parameters ($\Matrix{T}_{cr,prior}$) is derived from hand measurement. 
We set the prior uncertainty for the translation ($\sigma_t$) to 0.1 m along each axis, and the prior uncertainty for the rotation to ($\sigma_\theta$) 30 degrees.
The addition of this term stabilizes the estimation of the vertical translation between the radar and cameras, in particular.
After optimization, less than 1\% of the final cost value is due to the prior error term.

\begin{figure}[t] 
	\centering
	\begin{tikzpicture}
	\node[anchor=south west,inner sep=0] (image) at (0,0) {\includegraphics[width=\columnwidth]{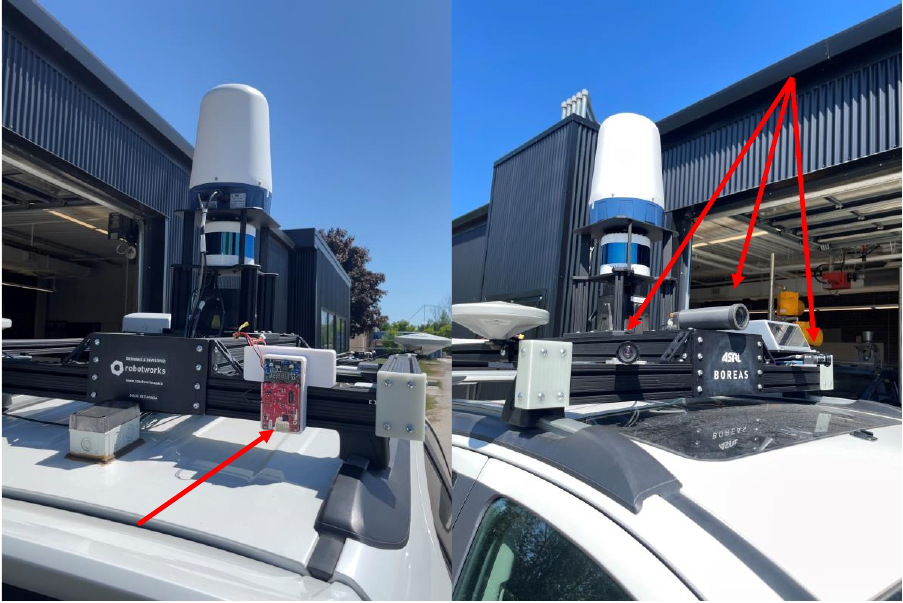}};
	\node (Radar) [rectangle,draw,fill=white, align=center] at (1.63, 0.48) {AWR1843BOOST\\3D mm-Wave Radar};
	\node (Camera) [rectangle,draw,fill=white, align=center] at (7.7, 5.42) {Point Grey\\GigE Cameras};
	\end{tikzpicture}
	\caption{Two views of the radar and camera mounting positions on the vehicle used in our experiments. The left image shows the mounting position of the TI radar. The right image shows the mounting positions of the three Point Grey cameras. The radar and the cameras do not share overlapping fields of view.}
	\label{fig:boreas_cameras}
\end{figure}

The mounting positions of the radar and three cameras on the car are shown in \Cref{fig:boreas_cameras}, and the corresponding fields of view are shown in \Cref{fig:boreas_fov}.
The first camera is positioned at the centre of the car and faces the direction of travel.
The other two cameras are placed to the left and right of the centre camera and point roughly 45 degrees left and right from the forward axis, respectively.
The 3D radar is more than one metre away from all of the cameras, facing towards the rear of the car, opposite the direction of travel.

We collected a total of nine datasets from the radar and the cameras (three datasets per camera) while driving two laps of a figure eight pattern.
Data collection took place in a sparse parking lot environment, where the radar and camera features were at a substantial distance from the vehicle.
We evaluated the accuracy of our estimated parameters by comparing the estimated distances between the centre camera and the two side cameras to the distances measured using a Leica Nova MS50 MultiStation.
This method of comparison was selected in part because camera-to-camera extrinsic calibration is difficult for camera pairs that have minimal field of view overlap.
Additionally, structural components of the car prevent direct measurement of the distance between the radar and cameras.
Each run of our spatiotemporal calibration algorithm produced an estimated extrinsic calibration, for a total of three sets of estimated extrinsic calibration parameters for each camera.
The transformations between the centre-to-left and -right cameras are computed by combining two radar extrinsic calibration estimates, which give a total of 18 camera-to-camera extrinsic calibration estimates (nine left and nine right).

\begin{figure}[b] 
	\centering
	\includegraphics[width = \columnwidth]{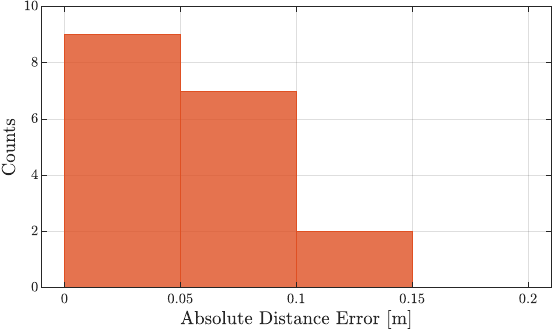}
	\caption{Results from the vehicle calibration experiment, where the radar and the cameras do not share overlapping fields of view. The distance error is the difference between the estimated and measured distances between the center-left and center-right cameras. The ground truth distance was determined using a Leica MultiStation.}
	\label{fig:boreas_results}
\end{figure}

Figure \ref{fig:boreas_results} shows the distribution of distance errors.
The majority of estimated extrinsic calibration parameters result in a camera-to-camera distance error of less than 5 cm, with two values that are greater than 10 cm.
This error is reasonable given the `chained' nature of the two radar-camera transforms.
The accuracy of the estimated camera-camera distance depends on the accuracy of the translation and rotation parameters for both transforms.
For this experiment, the radar-camera transforms have translation magnitudes greater than 1 m, so a small error in either estimated rotation results in a large distance error.
In turn, we expect the estimated rotation and translation parameters to be within 5$^\circ$ and 5 cm of the their true values.

\subsection{Calibration Environment}

Several notes are in order regarding environments that are suitable for calibration.
Although our algorithm does not require any retroreflective targets for the radar or a specific calibration pattern for the camera, there are nonetheless some limitations on where calibration can be performed.
To ensure accurate ego-velocity estimation, the calibration environment should contain, at minimum, four stationary landmarks, with more being better.
Also, to ensure accurate camera pose estimation using ORB-SLAM3, the scene should have sufficient lighting and visual texture.
As a result, calibration should generally not be performed in scenes with many moving targets, dim lighting, or inclement weather such as fog.
The accuracy of the camera pose estimates ultimately depends upon the specific SLAM algorithm that is chosen. 

\section{Conclusion}
\label{sec:conclusion}

In this paper, we described an algorithm that leverages radar ego-velocity estimates, unscaled camera pose measurements, and a continuous-time trajectory representation to perform targetless radar-to-camera spatiotemporal calibration.
We proved that the calibration problem is identifiable and determined the necessary  conditions for successful calibration.
Through simulation studies, we demonstrated that our algorithm is accurate, but can be sensitive to the amount of noise present in the radar range-rate measurements.
Further, we evaluated our algorithm in three different, real-world environments.
First, we showed, using data from a handheld sensor rig, that our approach can match the accuracy of target-based calibration methods.
Second, we presented results indicating that calibration can improve the localization performance of a hardware-triggered radar-camera-IMU system.
Finally, we established that our calibration framework can be applied to AV systems, where the radar and camera are mounted at a significant distance from each other and do not share overlapping fields of view.

There are several potential directions for future research. 
It would be valuable to develop a method to automatically determine the knot spacing required for the continuous-time spline representation.
Our calibration approach could naturally be extended to the multi-camera and multi-radar setting.
Other pairs of sensors could also be considered beyond radar-camera pairs, including radar-inertial sensor combinations, for example.

\begin{appendices}

\section{An Extension on the Observability of Radar-to-Camera Extrinsic Calibration from Wise et.\ al \cite{wise_continuous-time_2021}}
\label{appendix:observability}

In this appendix, we provide an extension to our earlier work in \cite{wise_continuous-time_2021} demonstrating that radar-to-camera spatial calibration (with a known temporal offset) is locally weakly observable.

\subsection{Notation for Nonlinear Observability Analysis}
\label{sec:app_notation}

In \Cref{sec:methodology}, we represent rotations as orthonormal matrices, which is convenient for the calibration problem but slightly more difficult to use for observability analyses.
In this section, we rely on unit quaternions to represent rotations, avoiding the need to use exponential functions.
We write a unit quaternion in `vector' form as
\begin{equation}
\Vector{q} = \bbm q_0 & \Vector{q}_v \ebm,
\end{equation}
with the scalar component $q_0$ and vector component $\Vector{q}_v$, such that $\Norm{\Vector{q}}_{2} =  1$. The conversion from unit quaternion to rotation matrix is given by the formula 
\begin{equation}
\Matrix{R}_{ab} =\Matrix{R}(\Vector{q}_{ab}) = (2 q_0^2 - 1)\Identity_3 + 2\Vector{q}_v\Transpose{\Vector{q}_v} + 2 q_0\,\Vector{q}_v^\wedge.
\end{equation} 
The quaternion kinematics are defined by
\begin{align}
\dot{\Vector{q}} = \frac{1}{2} \Matrix{\Xi}(\Vector{q})\,\Vector{\omega} =\frac{1}{2}\Matrix{\Omega}(\Vector{\omega})\Vector{q}, \\[2mm]
\Matrix{\Omega}(\Vector{\omega}) = \bbm 0 & -\Transpose{\Vector{\omega}} \\ \Vector{\omega} & -\Vector{\omega}^\wedge \ebm,\\[2mm]
\Matrix{\Xi}(\Vector{q}) = \bbm -\Transpose{\Vector{q}_v} \\ q_0 \Identity_3 +\Vector{q}_v^\wedge \ebm,
\end{align}
where $\Vector{\omega}$ is the angular velocity vector.
The following identity is useful for the observability proof,
\begin{equation} \label{eq:quat_der}
\begin{split}
\frac{\partial \Matrix{R}(\Vector{q})\Vector{p}}{\partial \Vector{q}} = & [\begin{matrix} (4 q_0\Identity_3 + 2 \Vector{q}_v^\wedge)\Vector{p} \end{matrix} \\ & \quad \begin{matrix} 2((\Transpose{\Vector{q}_v}\Vector{p})\Identity_3 + \Vector{q}_v\Transpose{\Vector{p}} - q_0\Vector{p}^\wedge) \end{matrix}],
\end{split}
\end{equation}
where $\Vector{p}$ is an arbitrary 3 $\times$ 1 vector.
If we transpose the rotation matrix on the left side of \Cref{eq:quat_der}, then the skew-symmetric terms on the right side change sign from positive to negative and vice versa.

\subsection{Local Weak Observability}
\label{sec:obs-proof}

Following the procedure outlined in \Cref{subsec:obs-rank}, we define the system equations, compute the respective Lie derivatives, and demonstrate that the nonlinear observability matrix has full column rank.
In the analysis here, the pose, velocity, and acceleration states of the radar-camera system are camera-centric (i.e., taken with respect to the camera and not the radar).
Since the camera-centric states can be used to determine the radar-centric states, this change does not affect the observability result.

Considering the camera frame $\CoordinateFrame{c}$, the radar frame $\CoordinateFrame{r}$, and the world frame $\CoordinateFrame{w}$, the state vector for the observability analysis is defined as
\begin{equation}
\begin{aligned}
\Vector{x} = &\left[\begin{matrix} \Transpose{\Vector{r}_w^{cw}} & \Transpose{\Vector{q}_{wc}} & \Transpose{\Vector{v}_w^{cw}} & \Transpose{\Vector{\omega}_c^{cw}} & \Transpose{\Vector{a}_w^{cw}} & \Transpose{\Vector{\alpha}_c^{cw}} \end{matrix}\right.\\
& \quad \Transpose{\left.\begin{matrix} \gamma & \Transpose{\Vector{r}_c^{rc}} & \Transpose{\Vector{q}_{cr}} \end{matrix}\right]},
\end{aligned}
\end{equation}
where $\Vector{r}$, $\Vector{v}$, and $\Vector{a}$ denote translation, linear velocity, and linear acceleration, respectively.
The vectors $\Vector{\omega}$ and $\Vector{\alpha}$ are the angular velocity and the angular acceleration, respectively.
Finally, $\gamma$ is the scale factor for the camera translation (for a monocular camera system).
The motion model for the system is
\begin{equation}
\dot{\Vector{x}} = \Vector{f}_0 (\Vector{x}) + \Vector{f}_1 (\Vector{x})= \bbm \Vector{0}_{3 \times 1} \\ 
\frac{1}{2}\Xi(\Vector{q}_{wc})\Vector{\omega}_{c}^{cw} \\
\Vector{0}_{3 \times 1}\\
\Vector{\alpha}_{c}^{cw}\\
\Vector{0}_{3 \times 1}\\
\Vector{0}_{3 \times 1}\\
0 \\
\Vector{0}_{3 \times 1}\\
\Vector{0}_{4 \times 1}\ebm 
+ \bbm \Vector{v}_w^{cw} \\ 
\Vector{0}_{4 \times 1}\\
\Vector{a}_w^{cw}\\
\Vector{0}_{3 \times 1}\\
\Vector{0}_{3 \times 1}\\
\Vector{0}_{3\times 1}\\
0\\
\Vector{0}_{3 \times 1}\\
\Vector{0}_{4 \times 1}\ebm.
\end{equation}
The measurement model equations for the (scaled) camera translation and rotation are, respectively,
\begin{equation}
\begin{aligned}
\Vector{h}_1 & = \gamma\,\Vector{r}_w^{cw}, \\
\Vector{h}_2 & = \Vector{q}_{wc}.
\end{aligned}
\end{equation}
Using the camera-centric model, it is possible to directly measure $\Vector{q}_{wc}$ and, following the result in \Cref{subsec:radar-cam}, to determine $\Vector{\omega}_c^{cw}$ and $\Vector{\alpha}_{c}^{cw}$.
Finally, the radar ego-velocity measurement equation is
\begin{equation}
\Vector{h}_3 = \Transpose{\Matrix{R}}(\Vector{q}_{cr})(\Transpose{\Matrix{R}}(\Vector{q}_{wc})\Vector{v}_w^{cw} + \Vector{\omega}_{c}^{cw\wedge}\Vector{r}_c^{rc}).
\end{equation}

The observability analysis requires the zeroth-, first-, and second-order Lie derivatives.
The zeroth-order Lie derivatives are
\begin{equation} \label{eq:zero-order-lie}
\begin{aligned}
\nabla L^0 \Vector{h}_1 & =  \bbm \gamma\Identity_{3} & \Vector{0}_{3 \times 16} & \Vector{r}_w^{cw} & \Matrix{0}_{3 \times 7}\ebm,\\
\nabla L^0 \Vector{h}_2 & = \bbm \Vector{0}_{4 \times 3} & \Identity_{4} & \Matrix{0}_{4 \times 20} \ebm, \\
\nabla L^0 \Vector{h}_3 & = [\begin{matrix} \Vector{0}_{3 \times 3} & \Matrix{A} & \Transpose{\Matrix{R}}(\Vector{q}_{cr})\Transpose{\Matrix{R}}(\Vector{q}_{wc}) \end{matrix} \\
& \quad \begin{matrix} -\Transpose{\Matrix{R}}(\Vector{q}_{cr})\Vector{r}_c^{rc\wedge} & \Vector{0}_{3 \times 7}  \end{matrix} \\
& \quad\quad \begin{matrix} \Transpose{\Matrix{R}}(\Vector{q}_{cr})\Vector{\omega}_{c}^{cw\wedge} & \Matrix{B} \end{matrix}],
\end{aligned}
\end{equation}
where
\begin{equation}
\begin{aligned}
\Matrix{A} & = \Transpose{\Matrix{R}}\!(\Vector{q}_{cr})\frac{\partial \Transpose{\Matrix{R}}(\Vector{q}_{wc})\Vector{v}_w^{cw}}{\partial \Vector{q}_{wc}},\\
\Matrix{B} & = \frac{\partial\!\Transpose{\Matrix{R}}\!(\Vector{q}_{cr})(\Transpose{\Matrix{R}}(\Vector{q}_{wc})\Vector{v}_w^{cw} + \Vector{\omega}_{c}^{cw\wedge}\Vector{r}_c^{rc})}{\partial \Vector{q}_{cr}}.
\end{aligned}
\end{equation}
The first-order Lie derivatives are
\begin{equation}
\label{eq:first-order-lie}
\begin{aligned}
\nabla L^1_{\Vector{f}_1} \Vector{h}_1 & = \bbm \Vector{0}_{3 \times 7} & \gamma\Identity_{3} & \Vector{0}_{3 \times 9} & \Vector{v}_w^{cw} & \Matrix{0}_{3 \times 7}\ebm ,\\
\nabla L^1_{\Vector{f}_0} \Vector{h}_2 & = [\begin{matrix} \Vector{0}_{4 \times 3} & \frac{1}{2}\Omega(\Vector{\omega}_{c}^{cw}) & \Vector{0}_{4\times 3} \end{matrix} \\
& \quad \begin{matrix} \frac{1}{2}\Xi(\Vector{q}_{wc}) & \Matrix{0}_{4 \times 14} \end{matrix}], \\
\nabla L^1_{\Vector{f}_0} \Vector{h}_3 & = [\begin{matrix} \Matrix{0}_{3 \times 3} & \Matrix{C} & \Matrix{D} & \Matrix{E} & \Matrix{0}_{3 \times 3} \end{matrix}\\
& \quad \begin{matrix} \Matrix{F} & \Matrix{0}_{3 \times 1} & \Transpose{\Matrix{R}}(\Vector{q}_{cr})\Vector{\alpha}_{c}^{cw\wedge} & \Matrix{G}\end{matrix}],\\
\nabla L^1_{\Vector{f}_1} \Vector{h}_3 & = [\begin{matrix} \Vector{0}_{3 \times 3} & \Matrix{H} & \Matrix{0}_{3\times 6} \end{matrix} \\
& \quad \begin{matrix}  \Transpose{\Matrix{R}}(\Vector{q}_{cr})\Transpose{\Matrix{R}}(\Vector{q}_{wc})& \Vector{0}_{3 \times 7} & \Matrix{L} \end{matrix}],
\end{aligned}
\end{equation}
where
\begin{equation}
\begin{aligned}
\Matrix{H} =  & \Transpose{\Matrix{R}}(\Vector{q}_{cr})\frac{\partial \Transpose{\Matrix{R}}(\Vector{q}_{wc})\Vector{a}_w^{cw}}{\partial \Vector{q}_{wc}}, \\
\Matrix{L} =  & \frac{\partial \Transpose{\Matrix{R}}(\Vector{q}_{cr})\Transpose{\Matrix{R}}(\Vector{q}_{wc})\Vector{a}_w^{cw}}{\partial \Vector{q}_{cr}}.
\end{aligned}
\end{equation}
We do not explicitly require the nonzero matrices, $\Matrix{C}$, $\Matrix{E}$, and $\Matrix{F}$, in \Cref{eq:first-order-lie} because the submatrix formed from the columns corresponding to the rotation states can be shown to be full rank.
The matrices $\Matrix{D}$ and $\Matrix{G}$ are required for the analysis, but we omit them here for brevity.
The second-order Lie derivatives are
\begin{equation} \label{eq:second-order-lie}
\begin{aligned}
\nabla L^2_{\Vector{f}_1} \Vector{h}_1 & = \bbm \Vector{0}_{3 \times 13} & \gamma\Identity_{3} & \Vector{0}_{3 \times 3} & \Vector{a}_w^{cw} & \Matrix{0}_{3 \times 7}\ebm, \\
\nabla L^2_{\Vector{f}_0} \Vector{h}_2 & = [\begin{matrix} \Vector{0}_{4 \times 3} & \frac{1}{4}(2\Matrix{\Omega}(\Vector{\alpha}_{c}^{cw}) - \Transpose{\Vector{\omega}_{c}^{cw}}\Vector{\omega}_{c}^{cw}\Identity_{4}) \end{matrix} \\
& \begin{matrix} \Vector{0}_{4\times 3} & -\frac{1}{2}\Vector{q}_{wc}\Transpose{\Vector{\omega}_{c}^{cw}} & \Vector{0}_{4\times 3} \end{matrix}\\
& \begin{matrix} \frac{1}{2}\Xi(\Vector{q}_{wc}) & \Matrix{0}_{4 \times 8} \end{matrix}].
\end{aligned}
\end{equation}

Stacking the gradients of the Lie derivatives, we arrive at the nonlinear observability matrix,
\begin{equation} \label{eq:obs-matrix}
\Matrix{O} = \bbm \nabla L^0\Vector{h}_1\\
\nabla L^1_{f_1}\Vector{h}_1\\ 
\nabla L^2_{f_1f_1}\Vector{h}_1\\ 
\nabla L^0\Vector{h}_2\\ 
\nabla L^1_{f_0}\Vector{h}_2\\ 
\nabla L^2_{f_0f_0}\Vector{h}_2\\ 
\nabla L^0\Vector{h}_3\\ 
\nabla L^1_{f_0}\Vector{h}_3\\ 
\nabla L^1_{f_1}\Vector{h}_3\ebm.
\end{equation}
This matrix can be shown to be full column rank (except when excitation of the system is insufficient), hence the system is locally weakly observable.

\end{appendices}

\bibliographystyle{IEEEtran}
\bibliography{refs}

\begin{thebibliography}{10}
\providecommand{\url}[1]{#1}
\csname url@samestyle\endcsname
\providecommand{\newblock}{\relax}
\providecommand{\bibinfo}[2]{#2}
\providecommand{\BIBentrySTDinterwordspacing}{\spaceskip=0pt\relax}
\providecommand{\BIBentryALTinterwordstretchfactor}{4}
\providecommand{\BIBentryALTinterwordspacing}{\spaceskip=\fontdimen2\font plus
\BIBentryALTinterwordstretchfactor\fontdimen3\font minus
  \fontdimen4\font\relax}
\providecommand{\BIBforeignlanguage}[2]{{%
\expandafter\ifx\csname l@#1\endcsname\relax
\typeout{** WARNING: IEEEtran.bst: No hyphenation pattern has been}%
\typeout{** loaded for the language `#1'. Using the pattern for}%
\typeout{** the default language instead.}%
\else
\language=\csname l@#1\endcsname
\fi
#2}}
\providecommand{\BIBdecl}{\relax}
\BIBdecl

\bibitem{lee_extrinsic_2020}
C.-L. Lee, Y.-H. Hsueh, C.-C. Wang, and W.-C. Lin, ``Extrinsic and {{Temporal
  Calibration}} of {{Automotive Radar}} and {{3D Lidar}},'' in \emph{2020
  {IEEE/RSJ} Intl. Conf. Intelligent Robots and Systems ({IROS})}, {Las Vegas,
  {NV}, {USA}}, Oct. 25, 2020--Jan. 24, 2021 2020, pp. 9976--9983.

\bibitem{persic_spatiotemporal_2021}
J.~Per\v{s}i\'{c}, L.~Petrovi\'{c}, I.~Markovi\'{c}, and I.~Petrovi\'{c},
  ``Spatiotemporal multisensor calibration via {Gaussian} processes moving
  target tracking,'' \emph{{IEEE} Trans. Robotics}, vol.~37, no.~5, pp.
  1401--1415, Mar. 2021.

\bibitem{richards_principle_2010}
M.~A. Richards, J.~A. Scheer, and W.~A. Holm, Eds., \emph{{Principles of Modern
  Radar: Basic Principles}}.\hskip 1em plus 0.5em minus 0.4em\relax Institution
  of Eng. and Technol., 2010, vol.~1.

\bibitem{wise_continuous-time_2021}
E.~Wise, J.~Per\v{s}i\'{c}, C.~Grebe, I.~Petrovi\'{c}, and J.~Kelly, ``A
  continuous-time approach for {3D} radar-to-camera extrinsic calibration,'' in
  \emph{2021 {IEEE} Intl. Conf. Robotics and Automation ({ICRA})}, Xi'an,
  China, May 30--Jun. 5 2021, pp. 13\,164--13\,170.

\bibitem{Stahoviak_Velocity_2019}
C.~C. Stahoviak, ``\BIBforeignlanguage{English}{An instantaneous {3D}
  ego-velocity measurement algorithm for frequency modulated continuous wave
  ({FMCW}) {Doppler} radar data},'' Master's thesis, University of Colorado at
  Boulder, 2019.

\bibitem{sugimoto_obstacle_2004}
S.~Sugimoto, H.~Tateda, H.~Takahashi, and M.~Okutomi, ``{Obstacle detection
  using millimeter-wave radar and its visualization on image sequence},'' in
  \emph{Int. Conf. Pattern Recognition (ICPR)}, Cambridge, England, Aug. 23--26
  2004, pp. 342--345.

\bibitem{Wang2011}
T.~Wang, N.~Zheng, J.~Xin, and Z.~Ma, ``{Integrating millimeter wave radar with
  a monocular vision sensor for on-road obstacle detection applications},''
  \emph{Sensors}, vol.~11, no.~9, pp. 8992--9008, Sep. 2011.

\bibitem{kim_data_2014}
D.~Y. Kim and M.~Jeon, ``\BIBforeignlanguage{en}{Data fusion of radar and image
  measurements for multi-object tracking via {{Kalman}} filtering},''
  \emph{\BIBforeignlanguage{en}{Information Sciences}}, vol. 278, pp. 641--652,
  Sep. 2014.

\bibitem{kim_radar_2018}
J.~Kim, D.~S. Han, and B.~Senouci, ``Radar and vision sensor fusion for object
  detection in autonomous vehicle surroundings,'' in \emph{2018 {{4th Int.
  Conf.}} {{Ubiquitous}} and {{Future Networks}} ({{ICUFN}})}, Prague, Czech
  Republic, Jul. 3--6 2018, pp. 76--78.

\bibitem{kim_comparative_2017}
T.~Kim, S.~Kim, E.~Lee, and M.~Park, ``Comparative analysis of {{RADAR}}-{{IR}}
  sensor fusion methods for object detection,'' in \emph{2017 17th {{Int.
  Conf.}} {{Control}}, {{Automation}} and {{Systems}} ({{ICCAS}})}, Jeju,
  Korea, Oct. 18--21 2017, pp. 1576--1580.

\bibitem{elnatour_radar_2015}
G.~El~Natour, O.~Ait~Aider, R.~Rouveure, F.~Berry, and P.~Faure, ``Radar and
  vision sensors calibration for outdoor {{3D}} reconstruction,'' in \emph{2015
  {{IEEE Int. Conf.}} {{Robotics}} and {{Automation}} ({{ICRA}})}, Seattle, WA,
  USA, May 25--30 2015, pp. 2084--2089.

\bibitem{Domhof2019_calibration}
J.~{Domhof}, J.~F.~P. {Kooij}, and D.~M. {Gavrila}, ``An extrinsic calibration
  tool for radar, camera and lidar,'' in \emph{2019 Int. Conf. Robotics and
  Automation (ICRA)}, Montr\'{e}al, Canada, May, 20--24 2019, pp. 8107--8113.

\bibitem{Persic2019_calibration}
J.~Per\v{s}i\'{c}, I.~Markovi\'{c}, and I.~Petrovi\'{c}, ``Extrinsic {6DoF}
  calibration of a radar--lidar--camera system enhanced by radar cross section
  estimates evaluation,'' \emph{Robotics and Autonomous Systems}, vol. 114, pp.
  217--230, Apr. 2019.

\bibitem{Oh2018_calibration}
J.~{Oh}, K.~{Kim}, M.~{Park}, and S.~{Kim}, ``A comparative study on
  camera-radar calibration methods,'' in \emph{2018 15th Int. Conf. Control,
  Automation, Robotics and Vision (ICARCV)}, Singapore, Nov. 18--21 2018, pp.
  1057--1062.

\bibitem{Knoll2019_calibration}
C.~{Sch{\"{o}}ller}, M.~{Schnettler}, A.~{Kr{\"{a}}mmer}, G.~{Hinz},
  M.~{Bakovic}, M.~{G{\"{u}}zet}, and A.~{Knoll}, ``Targetless rotational
  auto-calibration of radar and camera for intelligent transportation
  systems,'' in \emph{2019 IEEE Intelligent Transportation Systems Conf.
  (ITSC)}, Auckland, New Zealand, Oct. 27--30 2019, pp. 3934--3941.

\bibitem{Persic2020}
J.~Per\v{s}i\'{c}, L.~Petrovi\'{c}, I.~Markovi\'{c}, and I.~Petrovi\'{c},
  ``Online multi-sensor calibration based on moving object tracking,''
  \emph{Advanced Robotics}, vol.~35, no. 3--4, pp. 130--140, Sep. 2021.

\bibitem{heng_automatic_2020}
L.~Heng, ``Automatic targetless extrinsic calibration of multiple {3D} lidars
  and radars,'' in \emph{2020 {IEEE/RSJ} Intl. Conf. Intelligent Robots and
  Systems ({IROS})}, {Las Vegas, NV, USA}, Oct. 25, 2020--Jan. 24, 2021 2020,
  pp. 10\,669--10\,675.

\bibitem{Kellner2015_calibration}
D.~{Kellner}, M.~{Barjenbruch}, K.~{Dietmayer}, J.~{Klappstein}, and
  J.~{Dickmann}, ``Joint radar alignment and odometry calibration,'' in
  \emph{2015 18th Int. Conf. Information Fusion (FUSION)}, Washington, DC, USA,
  Jul. 6--9 2015, pp. 366--374.

\bibitem{doer_radar_2020}
C.~Doer and G.~F. Trommer, ``Radar inertial odometry with online calibration,''
  in \emph{2020 European Navigation Conf. ({ENC})}, Nov. 23--24 2020, pp.
  1--10.

\bibitem{2016_Rehder_General}
J.~Rehder, R.~Siegwart, and P.~Furgale, ``A general approach to spatiotemporal
  calibration in multisensor systems,'' \emph{{IEEE} Trans. Robotics}, vol.~32,
  no.~2, pp. 383--398, Apr. 2016.

\bibitem{barfoot2017state}
T.~D. Barfoot, \emph{State estimation for robotics}.\hskip 1em plus 0.5em minus
  0.4em\relax Cambridge, UK: Cambridge Univ. Press, 2017.

\bibitem{Sommer_Efficient_2019}
C.~{Sommer}, V.~{Usenko}, D.~{Schubert}, N.~{Demmel}, and D.~{Cremers},
  ``Efficient derivative computation for cumulative {B}-splines on {Lie}
  groups,'' in \emph{2020 IEEE/CVF Conf. Computer Vision and Pattern
  Recognition (CVPR)}, Jun. 14--19 2020, pp. 11\,145--11\,153.

\bibitem{1978_Boor_Splines}
C.~de~Boor, \emph{A Practical Guide to Splines}, ser. Applied Mathematical
  Sciences.\hskip 1em plus 0.5em minus 0.4em\relax Springer-Verlag, Jan. 1978,
  vol.~27.

\bibitem{1998_Qin_Splines}
K.~Qin, ``General matrix representations for b-splines,'' in \emph{6th Pacific
  Conf. on Computer Graphics and Applications}, Singapore, Oct. 26--29 1998,
  pp. 37--43.

\bibitem{doer_reve_2020}
C.~Doer and G.~F. Trommer, ``An {EKF} based approach to radar inertial
  odometry,'' in \emph{2020 {IEEE} Intl. Conf. Multisensor Fusion and
  Integration for Intelligent Systems ({MFI})}, Karlsruhe, Germany, Sep. 14--16
  2020, pp. 152--159.

\bibitem{chiuso2002_structure}
A.~{Chiuso}, P.~{Favaro}, {Hailin Jin}, and S.~{Soatto}, ``Structure from
  motion causally integrated over time,'' \emph{IEEE Trans. Pattern Analysis
  and Machine Intelligence}, vol.~24, no.~4, pp. 523--535, Apr. 2002.

\bibitem{ceres-solver}
\BIBentryALTinterwordspacing
S.~Agarwal \emph{et~al.}, \emph{Ceres Solver}. [Online]. Available:
  \url{http://ceres-solver.org}
\BIBentrySTDinterwordspacing

\bibitem{2014_Li_Online}
M.~Li and A.~I. Mourikis, ``Online temporal calibration for camera-{IMU}
  systems: Theory and algorithms,'' \emph{Intl. J. Robotics Research}, vol.~33,
  no.~7, pp. 947--964, 2014.

\bibitem{2015_Hewitt_Towards}
R.~A. Hewitt and J.~A. Marshall, ``Towards intensity-augmented {SLAM} with
  {LiDAR} and {ToF} sensors,'' in \emph{Proc. {IEEE/RSJ} Intl. Conf.
  Intelligent Robots and Systems {(IROS)}}, Hamburg, Germany, September/October
  2015, pp. 1957--1961.

\bibitem{hermann_nonlinear_1977}
R.~Hermann and A.~Krener, ``Nonlinear controllability and observability,''
  \emph{IEEE Trans. Automatic Control}, vol.~22, no.~5, pp. 728--740, Oct.
  1977.

\bibitem{2021_Kelly_Question}
J.~Kelly, C.~Grebe, and M.~Giamou, ``A question of time: Revisiting the use of
  recursive filtering for temporal calibration of multisensor systems,'' in
  \emph{Proc. {IEEE} Intl. Conf. Multisensor Fusion and Integration {(MFI)}},
  Karlsruhe, Germany, 2021.

\bibitem{doer_rrxio_2021}
C.~Doer and G.~F. Trommer, ``Radar visual inertial odometry and radar thermal
  inertial odometry: Robust navigation even in challenging visual conditions,''
  in \emph{2021 {IEEE/RSJ} Intl. Conf. Intelligent Robots and Systems
  ({IROS})}, Prague, Czech Republic, Sep. 27 -- Oct. 1 2021, pp. 331--338.

\bibitem{2023_burnett_boreas}
K.~Burnett, D.~J. Yoon, Y.~Wu, A.~Z. Li, H.~Zhang, S.~Lu, J.~Qian, W.-K. Tseng,
  A.~Lambert, K.~Y. Leung, A.~P. Schoellig, and T.~D. Barfoot, ``Boreas: A
  multi-season autonomous driving dataset,'' \emph{The International Journal of
  Robotics Research}, vol.~42, pp. 33--42, 2023.

\bibitem{AWR1843BOOST}
\BIBentryALTinterwordspacing
{Texas Instruments}, \emph{xWR1843 Evaluation Module (xWR1843BOOST) Single-Chip
  mmWave Sensing Solution}, May 2020. [Online]. Available:
  \url{https://www.ti.com/lit/ug/spruim4b/spruim4b.pdf}
\BIBentrySTDinterwordspacing

\bibitem{IWR6843AOP}
\BIBentryALTinterwordspacing
------, \emph{IWR6843AOP Single-Chip 60- to 64-GHz mmWave Sensor
  Antennas-On-Package (AOP)}, July 2022. [Online]. Available:
  \url{https://www.ti.com/lit/ds/symlink/iwr6843aop.pdf}
\BIBentrySTDinterwordspacing

\bibitem{campos_orb3_2021}
C.~Campos, R.~Elvira, J.~J.~G. Rodr'{i}guez, J.~M.~M. Montiel, and J.~D.
  Tard\'{o}s, ``{ORB-SLAM3}: An accurate open-source library for visual,
  visual–inertial, and multimap {SLAM},'' \emph{{IEEE} Trans. Robotics},
  vol.~37, no.~6, pp. 1874--1890, May 2021.

\bibitem{olson_tags_2011}
E.~Olson, ``{AprilTag}: A robust and flexible visual fiducial system,'' in
  \emph{2011 {IEEE} Intl. Conf. Robotics and Automation ({ICRA})}, Shanghai,
  China, May 9--13 2011, pp. 3400--3407.

\end{thebibliography}

\begin{IEEEbiography}[{\includegraphics[width=1in,height=1.25in,clip,keepaspectratio]{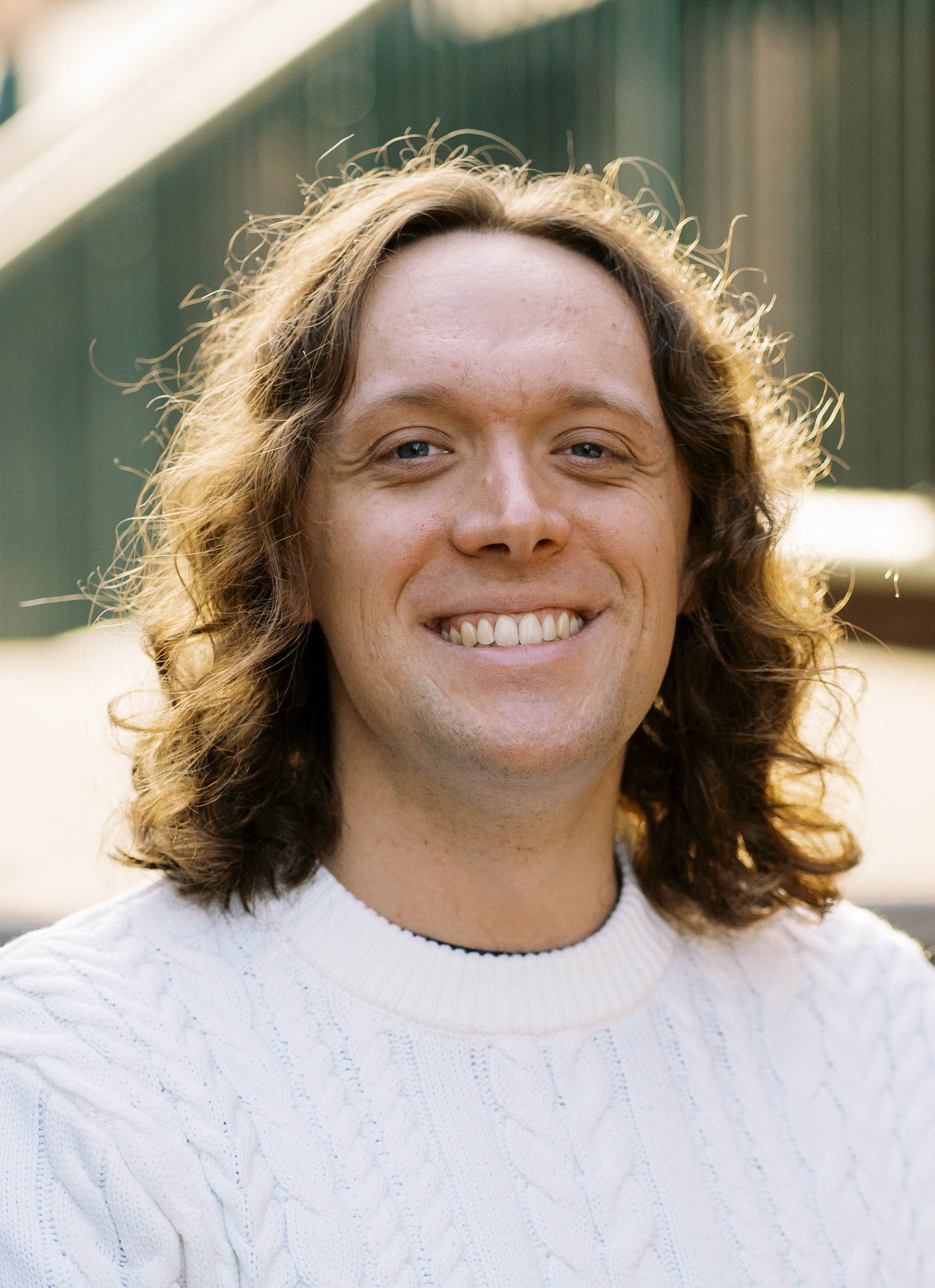}}]{Emmett Wise} received his Bachelor of Applied Science in engineering physics from Queens University, Kingston, Canada, in 2016. After graduation, he worked to automate the manufacturing of nanoscale coatings at 3M Canada. He is currently a Ph.D.\ Candidate in the Space and Terrestrial Autonomous Robotics (STARS) Laboratory at the Institute for Aerospace Studies, Toronto, Canada. His research interests include perception, calibration, and state estimation.
\end{IEEEbiography}
\begin{IEEEbiography}[{\includegraphics[width=1in,height=1.25in,clip,keepaspectratio]{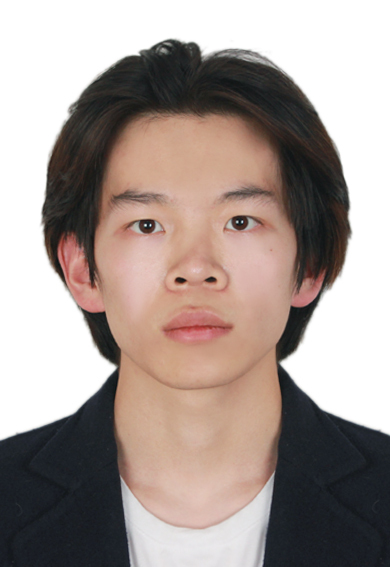}}]{Qilong (Jerry) Cheng} is currently pursuing his Master of Engineering degree in electrical and computer engineering, having completed a Bachelor’s in Mechanical Engineering at the University of Toronto. He was a developer at Autodesk from 2019 to 2020. From 2020 to 2021, he made design patent contributions to a high pressure spray nozzle design for the China State Shipbuilding Corporation. He is currently a graduate research student in the Space and Terrestrial Autonomous Robotics (STARS) Laboratory at the University of Toronto, focusing on calibration and state estimation. 
\end{IEEEbiography}
\begin{IEEEbiography}[{\includegraphics[width=1in,height=1.25in,clip,keepaspectratio]{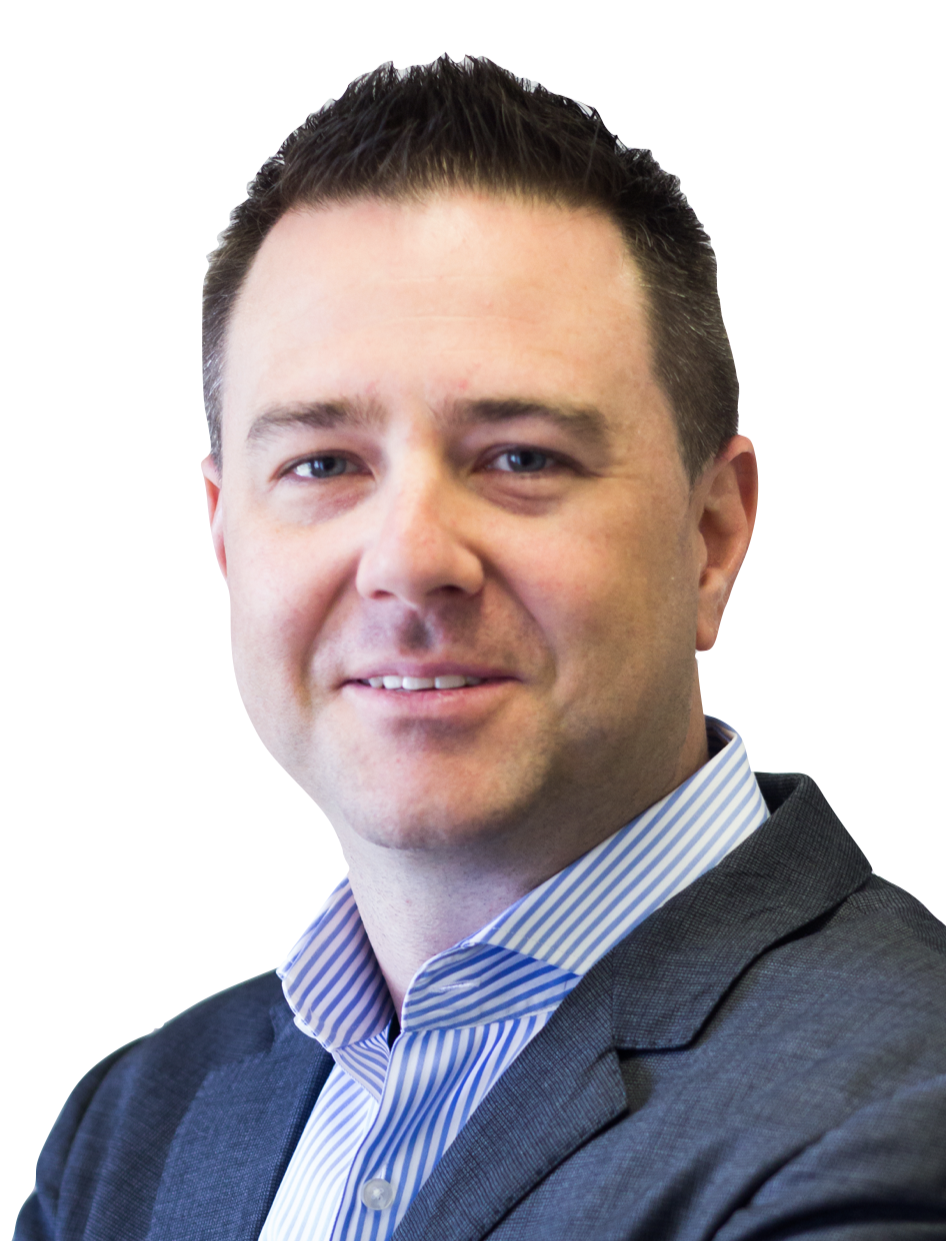}}]{Jonathan Kelly} received the Ph.D.\ degree in Computer Science from the University of Southern California, Los Angeles, USA, in 2011. From 2011 to 2013 he was a postdoctoral associate in the Computer Science and Artificial Intelligence Laboratory at the Massachusetts Institute of Technology, Cambridge, USA. He is currently an associate professor and director of the Space and Terrestrial Autonomous Robotic Systems (STARS) Laboratory at the University of Toronto Institute for Aerospace Studies, Toronto, Canada. Prof. Kelly holds the Tier II Canada Research Chair in Collaborative Robotics. His research interests include perception, planning, and learning for interactive robotic systems.
\end{IEEEbiography}

\end{document}